\documentclass[letterpaper]{article} 
\usepackage{aaai25}  
\usepackage{times}  
\usepackage{helvet}  
\usepackage{courier}  
\usepackage[hyphens]{url}  
\usepackage{graphicx} 
\usepackage{natbib}  
\usepackage{caption} 
\usepackage{algorithm}
\usepackage{algorithmic}

\usepackage{makecell}
\usepackage{booktabs}
\usepackage{amssymb}
\usepackage{multirow}
\usepackage[table]{xcolor}
\usepackage{pdfpages}
\definecolor{darkblue}{HTML}{1D305F}

\usepackage{newfloat}
\usepackage{listings}
\DeclareCaptionStyle{ruled}{labelfont=normalfont,labelsep=colon,strut=off} 
\floatstyle{ruled}
\newfloat{listing}{tb}{lst}{}
\floatname{listing}{Listing}
\title{Advancing Comprehensive Aesthetic Insight with Multi-Scale Text-Guided Self-Supervised Learning}
\author{
    Yuti Liu\equalcontrib, Shice Liu\equalcontrib, Junyuan Gao, Pengtao Jiang, Hao Zhang, Jinwei Chen, Bo Li\thanks{This author is the corresponding author.}
}
\affiliations{
    vivo Mobile Communication Co., Ltd\\
    Shanghai, China\\
    \{kira66arik\equalcontrib, junyuangao577\}@gmail.com; \{liushice\equalcontrib, pt.jiang, haozhang, jinwei.chen, libra$^{\dag}$\}@vivo.com

}

\begin{document}

\maketitle

\begin{abstract}
Image Aesthetic Assessment (IAA) is a vital and intricate task that entails analyzing and assessing an image's aesthetic values, and identifying its highlights and areas for improvement. Traditional methods of IAA often concentrate on a single aesthetic task and suffer from inadequate labeled datasets, thus impairing in-depth aesthetic comprehension. Despite efforts to overcome this challenge through the application of Multi-modal Large Language Models (MLLMs), such models remain underdeveloped for IAA purposes. To address this, we propose a comprehensive aesthetic MLLM capable of nuanced aesthetic insight. Central to our approach is an innovative multi-scale text-guided self-supervised learning technique. This technique features a multi-scale feature alignment module and capitalizes on a wealth of unlabeled data in a self-supervised manner to structurally and functionally enhance aesthetic ability. The empirical evidence indicates that accompanied with extensive instruct-tuning, our model sets new state-of-the-art benchmarks across multiple tasks, including aesthetic scoring, aesthetic commenting, and personalized image aesthetic assessment. Remarkably, it also demonstrates zero-shot learning capabilities in the emerging task of aesthetic suggesting. Furthermore, for personalized image aesthetic assessment, we harness the potential of in-context learning and showcase its inherent advantages.
\end{abstract}

%

\begin{figure*}[t]
    \centering
    \includegraphics[width=1.0\textwidth]{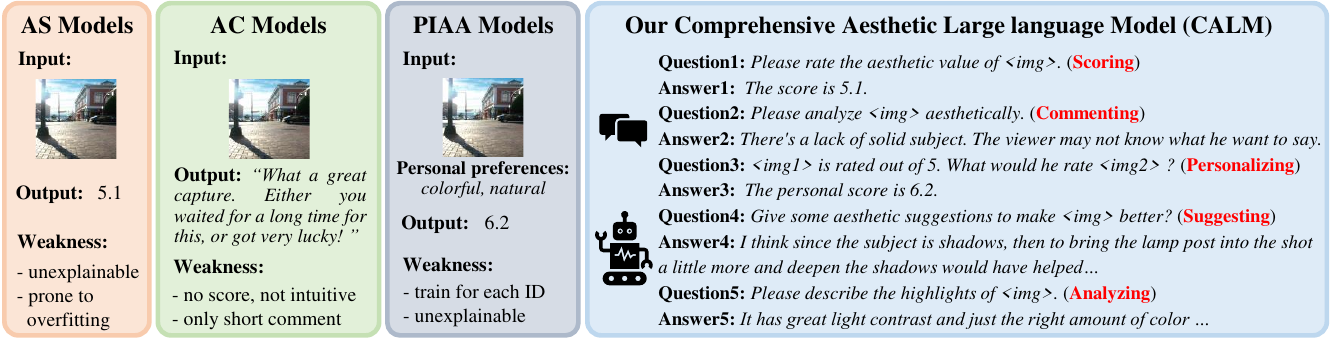}
    \caption{The functional comparison of our proposed CALM and other IAA methods.}
    \label{fig:first_page_compare}
\end{figure*}

\section{Introduction}
\label{sec:intro}

As artificial intelligence evolves, there's a growing demand for agents to mimic human perception and exhibit emotional responses to their surroundings. IAA emerges as a key area within this scope, and gauges images' aesthetic appeal akin to human judgment. Its complexity lies in its subjectivity, governed by factors like photographic subjects and personal experiences, which makes IAA a challenging endeavor.

In the last decade, IAA has been concretized into a variety of tasks. EAT \cite{he2023eat} predicts aesthetics based on a single human-assigned score per image---a task known as Aesthetic Scoring (AS). Meanwhile, CWS \cite{ghosal2019aesthetic} assesses an image's aesthetic appeal directly through language, which is referred to Aesthetic Commenting (AC). Recently, Personalized Image Aesthetic Assessment (PIAA) has emerged as a burgeoning field, which aims to predict an individual's aesthetic preferences based on his historical image scoring. While effective in certain scenarios, approaches focusing on a single task often fail to address linkages between different tasks, suffering from overfitting to specific tasks. This realization has inspired us to prioritize holistic aesthetic analysis and comprehension in our research efforts.

Recently, MLLMs have demonstrated strong comprehension and reasoning abilities across various domains. Models such as VILA \cite{ke2023vila}, Q-Align \cite{wu2023q}, and UNIAA \cite{zhou2024uniaa} have also attempted to utilize MLLMs for IAA to compensate for perceptual and reasoning processes. However, two major obstacles limit their effectiveness. First, these models rely solely on semantic features, neglecting a wealth of valuable aesthetic information. Second, despite efforts by Q-Align and UNIAA to construct aesthetic question-answer pairs for enhancement, the scarcity of labeled data and the presence of potentially mislabeled data continue to restrict their performance. Consequently, integrating comprehensive aesthetic information into MLLMs and developing a refined learning strategy to accurately leverage massive image data are essential.

In this paper, we propose Comprehensive Aesthetic Large language Model (CALM) which excels in various IAA tasks and demonstrates deep aesthetic comprehension and analytical skills in dialogues. Fig. \ref{fig:first_page_compare} illustrates the functional differences between CALM and other IAA models.

Inspired by popular MLLMs, CALM incorporates a visual encoder, a Multi-scale Feature Alignment Module (MFAM) and a Large Language Model (LLM). Recognizing that mainstream visual encoders and LLMs excel at feature extraction and language expression, we have focused our efforts on the MFAM to ensure that the subsequent LLM can fully leverage a broader spectrum of aesthetic information provided by the visual encoder. To achieve this, we introduce a multi-scale text-guided self-supervised learning technique.

Specifically, the MFAM is designed to structurally access aesthetic features at multiple levels, while text-guided self-supervised learning enables the MFAM to benefit from unlabeled data. Unlike previous aesthetic self-supervised approaches that rely on score pseudo-labels, our method uses attribute-related textual pseudo-labels. This change ensures accurate learning and simplifies the integration of pseudo-labels when superimposing multiple augmentations on a single image. Additionally, we utilize a wider range of image augmentations, from low-level to high-level, to guarantee that more aesthetic elements are captured and learned.

To enhance holistic aesthetic insight, we developed various instruct-tuning techniques to adapt CALM to common aesthetic tasks, ultimately outperforming other approaches in AS, AC, and PIAA tasks. Moreover, CALM achieves comparable PIAA results through in-context learning at runtime, establishing a new paradigm for PIAA. Additionally, we are the first to define the aesthetic suggesting task, and CALM's zero-shot success in this task demonstrates its ability to grasp and comprehend aesthetic principles effectively.

The contributions of our work are concluded as follows:

$\diamondsuit$ We propose CALM, a cutting-edge multi-modal large language model specialized in comprehending image aesthetics. Our extensive experiments demonstrate that CALM sets a new benchmark for AS, AC, and PIAA tasks.

$\diamondsuit$ We have pioneered a multi-scale text-guided self-sup-ervised learning technique that not only ensures multi-scale perception for MLLMs, but also effectively and efficiently leverages abundant unlabeled images for enhancement.

$\diamondsuit$ The remarkable zero-shot capabilities of CALM are explored, particularly in in-context PIAA and providing aesthetic suggestions. These capabilities demonstrate CALM's comprehensive aesthetic insight and analytical prowess.

\section{Related Work}

\textbf{Image Aesthetic Assessment} involves algorithms that measure the visual appeal of images. Initially, convolutional neural networks (CNN) and transformers have been leveraged to refine aesthetic score predictions, such as TANet \cite{he2022rethinking}, ResNext \cite{hou2022interaction}, DAT \cite{xia2022vision} and MaxViT \cite{tu2022maxvit}. In order to regulate aesthetic features to refine scoring, Comm \cite{niu2022comment} and AesCLIP \cite{sheng2023aesclip} harness textual data and CLIP \cite{radford2021learning}, respectively. Besides, language generation models for AC task have also emerged, such as Yeo \cite{yeo2021generating}. Moreover, realizing the importance of personal tastes, models and the FLICKR-AES dataset \cite{ren2017personalized} for PIAA are gaining traction. However, previous methods usually concentrate on a single aesthetic task so that they can barely really understand aesthetics.

\begin{figure*}[t]
    \centering
    \includegraphics[width=1.0\textwidth]{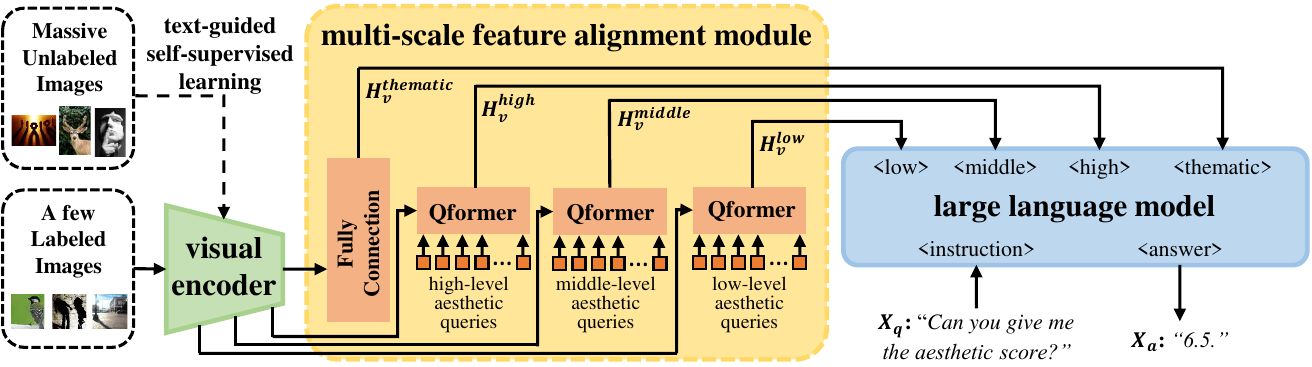}
    \caption{The proposed CALM includes a visual encoder, a multi-scale feature alignment module and a large language model.}
    \label{fig:framework}
\end{figure*}

\textbf{Multi-modal Large Language Models} achieve image content analysis by integrating visual features in LLMs. LLaVA-1.5 \cite{liu2023improved} and mPLUG-Owl2 \cite{ye2024mplug} have showcased impressive image reasoning skills. In the realm of IAA, VILA employs CoCa \cite{yu2022coca} to explore zero-shot aesthetic judgement, while Q-ALIGN directly utilizes the original mPLUG-Owl2. UNIAA leverages ChatGPT to generate comments to fine-tune LLaVA-1.5. However, these methods do not modify the pre-existing MLLMs and rely on a limited number of constructed data, which may prevent a comprehensive aesthetic understanding. Consequently, it is vital to improve the structural design and functional learning for deeper aesthetic comprehension.

\textbf{Multi-scale Aesthetic Perception} is a key approach for promoting IAA. \cite{chen2020adaptive} combined multi-level spatial features and employed adaptive dilated CNNs, while Comm designed a module to process multi-scale features. EAT and ICAA \cite{he2023thinking} incorporated interest points and delegate transformers, aligning better on specific scales. Drawing on these observations, we develop a technique for MLLMs that harnesses multi-scale features effectively.

\textbf{Self-supervised Learning} seeks to leverage large quantities of unlabeled data and artificially assigned pseudo-labels to enhance models' generalization. In IAA, where expert annotation is often costly, self-supervised methods are particularly prevalent.  \cite{sheng2020revisiting,pfister2021self} intuitively assigned lower aesthetic scores to augmented images for contrastive learning, generating score pseudo-labels. However, due to the still ambiguous factors influencing aesthetics and cases where depth-of-field blur can enhance aesthetic appeal, these methods risk producing inaccurate pseudo-scores. Moreover, these methods primarily focus on low-level data augmentations and require separate classifiers to regress scores, limiting their effectiveness.

\section{Methodology}

\subsection{The Architecture of CALM}
\label{subsec:preliminary}

As represented in Fig. \ref{fig:framework}, CALM is composed of three principal elements: a visual encoder $g(\cdot)$ transforming an image $\mathrm{X_v}$ into a sequence of visual tokens $\mathrm{Z_v}=g(X_v)$; an MFAM $W(\cdot)$ converting visual tokens $\mathrm{Z_v}$ into vision-language tokens $\mathrm{H_v}=W(\mathrm{Z_v})$; an LLM $f(\cdot)$ that receives the vision-language tokens $\mathrm{H_v}$ and user instructions $\mathrm{X_q}$ to produce the relevant language responses $\mathrm{X_a}=f(\mathrm{H_v},\mathrm{X_q})$.

Referring to most MLLMs, we employ the open-sourced ViT-L/14 as $g(\cdot)$ and Vicuna-7B \cite{chiang2023vicuna} as $f(\cdot)$ without any modification. For our proposed MFAM, we detail its structural design in Sec. \ref{subsec:multiscale} and its functional promotion via text-guided self-supervised learning in Sec. \ref{subsec:textguided}. Subsequently, we show how CALM simultaneously addresses various IAA tasks through two-stage instruct tuning in Sec. \ref{subsec:comprehensive}. What's more, only regression loss is employed to reduce the gap between $\mathrm{X_a}$ and the ground truth $\mathrm{X_{gt}}$.

\subsection{Multi-scale Feature Alignment Module}
\label{subsec:multiscale}

\cite{jin2019aesthetic} have revealed that image clarity and color schemes are encoded in lower-level features, while composition and impression requires higher-level features for interpretation. Although multi-scale features have been broadly explored in IAA, MLLMs, which typically process tokens from the last several layers of the visual encoder, lacks a structural basis for handling multi-scale features. Hence, we design the MFAM to emphasize multi-scale information.

We define four levels based on their positions in $g(\cdot)$, from shallow to deep sequentially named as low-, middle-, high-, and thematic-level. To preserve the original reasoning ability, we utilize fully connection to yield thematic-level features $\mathrm{H_v^{thematic}}\in \mathcal{R}^{N_v\times d_l}$, where $N_v$ and $d_l$ are the number of vision tokens and the dimension of language tokens, respectively. And then, three two-layer Qformers \cite{li2023blip} are introduced, which use cross attentions to make learnable queries pinpoint aesthetic features at the targeted levels. With $g(\cdot)$ offering 24 hidden state layers, we strategically tap into the 4th, 12th, and 24th layers to compute low-level features $\mathrm{H_v^{low}}\in \mathcal{R}^{N_{low}\times d_l}$, middle-level features $\mathrm{H_v^{middle}}\in \mathcal{R}^{N_{middle}\times d_l}$, and high-level features $\mathrm{H_v^{high}}\in \mathcal{R}^{N_{high}\times d_l}$, where $N_{low}$, $N_{middle}$ and $N_{high}$ denote the number of learnable queries at each level. The design of MFAM makes it effective and efficient to capture key aesthetic features, considering that the number of queries is much smaller than that of visual tokens.

\subsection{Text-guided Self-supervised Learning}
\label{subsec:textguided}

For the purpose of effectively unlocking the potential of abundant unlabeled image data to accurately enhance aesthetic perception, we propose text-guided self-supervised learning, which offers the following three advantages.

Firstly, we use accurate attribute pseudo-labels to replace flawed score pseudo-labels for self-supervision. Concretely, we introduce various image augmentation algorithms targeting attributes mentioned in \cite{jin2019aesthetic}, such as color and subject. During training, unlabeled images are randomly augmented in certain attributes and assigned the corresponding attribute pseudo-labels. For instance, if an image is blurred, its attribute pseudo-label is "the blurred image".

Secondly, we leverage a broader spectrum of data augmentations compared to previous aesthetic self-supervised methods. These include low-level augmentations such as blurring and brightness adjustments, as well as high-level augmentations like cropping and masking significant objects. Details of all augmentations and their corresponding pseudo-labels can be found in the \textit{Appendix A}. Subsequent experiments confirm that these image augmentations significantly enhance aesthetic insight.

\begin{figure*}[t]
    \centering
    \includegraphics[width=\textwidth]{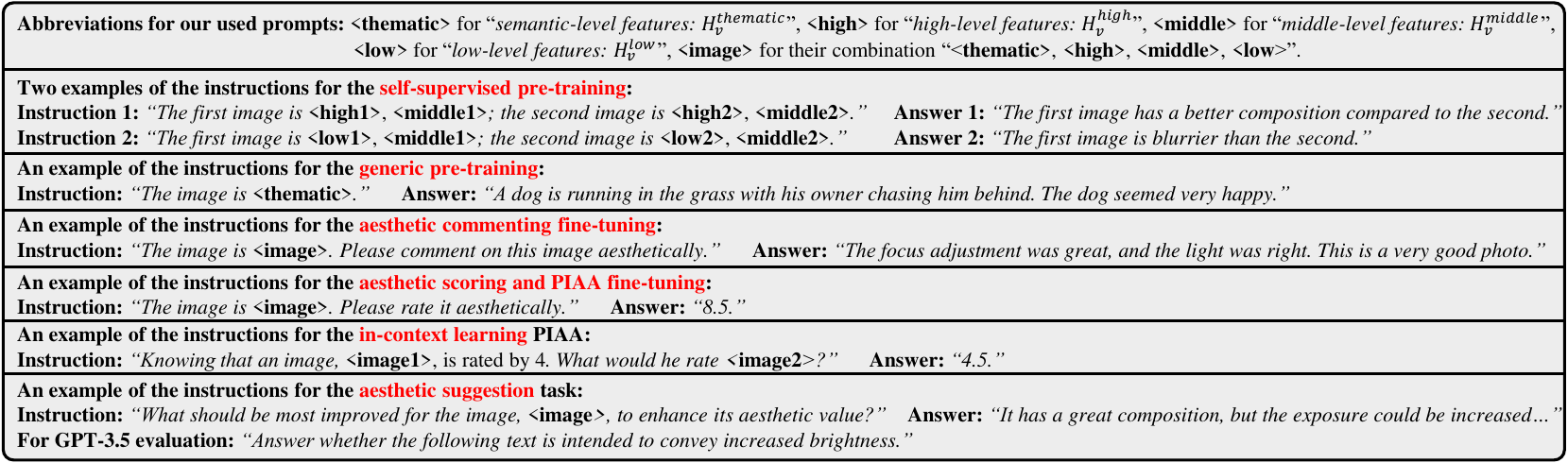}
    \caption{Some instruction examples utilized throughout the entire training process and across various tasks.}
    \label{fig:instruct}
\end{figure*}

Thirdly, we employ GPT-3.5 to generate various textual contrastive pseudo-labels, which eliminates the need for specialized classifiers in \cite{jin2019aesthetic}. Two examples are provided in the self-supervised pre-training part in Fig. \ref{fig:instruct}. Additionally, multiple augmentations can be applied simultaneously with their textual pseudo-labels conveniently spliced into a cohesive target, increasing both the data volume and the variety of contrastive learning. For instance, if an image is blurred and added noise, the pseudo-label would be, "The first image is blurrier and noisier than the second".

\begin{figure}[t]
    \centering
    \includegraphics[width=\columnwidth]{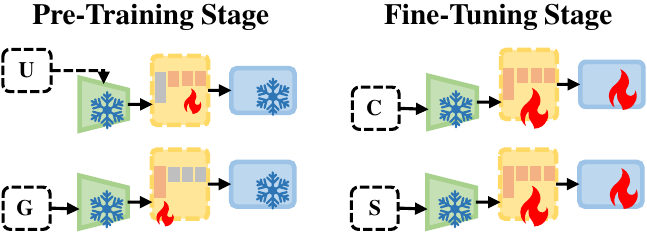}
    \caption{The two-stage training procedure. The pre-training stage focuses solely on the MFAM, while the fine-tuning stage also refines the LLM. The datasets are: unlabeled images (U), generic image-text pairs (G), aesthetic image-comment pairs (C), and aesthetic image-score pairs (S).}
    \label{fig:four_phases}
\end{figure}

\subsection{Comprehensive Aesthetic Assessment}
\label{subsec:comprehensive}

To achieve comprehensive aesthetic insight, we employ two-stage instruct tuning to adapt CALM to various aesthetic tasks, such as AS, AC, and PIAA. Specific instruction examples are shown in Fig. \ref{fig:instruct}. The complete training cycle, illustrated in Fig. \ref{fig:four_phases}, encompasses pre-training and fine-tuning.

The pre-training stage consists of two parts that can be launched simultaneously. \textbf{Self-Supervised Pre-Training} encourages the three Qformers in the MFAM to learn aesthetic attributes in a self-supervised manner, utilizing unlabeled images from diverse sources, including AVA \cite{murray2012ava}, AADB \cite{kong2016photo}, EVA \cite{kang2020eva}, ICAA, PCCD \cite{chang2017aesthetic}, pexels \cite{pfister2021self}, SPAQ \cite{fang2020perceptual} and TAD66K \cite{he2022rethinking}. To refine the learning process, augmentations on quality and color are designed to optimize $\mathrm{H_v^{middle}}$ and $\mathrm{H_v^{low}}$, while those on topics and composition benefit $\mathrm{H_v^{middle}}$ and $\mathrm{H_v^{high}}$. \textbf{Generic Pre-Training} focuses on training the fully connection to align $\mathrm{H_v^{thematic}}$, considering the value of generic knowledge for IAA \cite{ke2023vila}. The training data comprises a 558K subset of LAION-CC-SBU \cite{schuhmann2022laion,changpinyo2021conceptual,saleh2015large} and ShareGPT4V \cite{chen2023sharegpt4v}.

The fine-tuning stage consists of two task-specific processes that fine-tune MFAM and LLM concurrently. \textbf{Aesthetic Commenting Fine-Tuning} uses the AVA-Captions dataset \cite{ghosal2019aesthetic} to address AC task. \textbf{Aesthetic Scoring and PIAA Fine-Tuning} follows the aesthetic commenting fine-tuning, based on the insight from VILA that mastering AC can bolster effectiveness in AS. We use the AVA dataset for AS and the FLICKR-AES dataset for PIAA.

Having progressed through the two training stages, we are thrilled to find that CALM exhibits a strong aesthetic insight, primarily in the ability to accomplish some zero-shot activities such as giving aesthetic suggestions and conducting in-context PIAA. We highlight an examples of this in Fig. \ref{fig:instruct} and share more detailed experimental results later.

\section{Experiments}

\subsection{Experimental Settings}

\textbf{Datasets.} The AVA dataset comprises over 250,000 images with scores rated by users on the DPChallenge website. We used the official split, designating 19,928 images as the test set and the remainder for training. The AVA-Captions dataset contains approximately 230,000 images, each with an average of 5 user comments. To prevent data leakage, images from the AVA test set are excluded from AVA-Captions training, resulting in 210,000 images for training and 9,361 for testing. FLICKR-AES includes 35,263 images rated by 173 annotators in the training set and 4,737 images evaluated by 37 annotators in the test set, along with user identifications. Additionally, during the pre-training stage, around 460,000 unlabeled images and approximately 660,000 generic image-text pairs were utilized. Notably, we can further expand the unlabeled images as needed.

\textbf{Implementation Details.} The input resolution for $g(\cdot)$ is 224, and $N_v=256$ visual tokens are processed, each with a dimension of $d_v=1024$. In subsequent experiments, we set $N_{low}$, $N_{middle}$ and $N_{high}$ to 32. The dimension of language tokens is $d_l=4096$. Training was conducted on eight 80GB A100 GPUs, utilizing the Adam optimizer. The peak learning rate was set to 1e-3 for the pre-training stage, and 2.5e-5 and 7e-5 for the two processes in the fine-tuning stage, respectively. Both stages commenced with a linear warm-up, followed by a cosine annealing schedule, with durations of 5 hours and 16.5 hours, respectively.

\begin{table*}[ht!]
  \caption{The comparison results on the AS task. The methods with \colorbox{darkblue!30}{dark blue} marks use extra constructed aesthetic QA data.}
  \label{tab:avascore}
  \centering
  \belowrulesep=0pt
  \aboverulesep=0pt
  \resizebox{\textwidth}{!}{
    \setlength{\tabcolsep}{2pt}
    \begin{tabular}{c|ccc|ccc|ccc|ccc|cccc|ccc}
    \toprule
    \multirow{2}[2]{*}{Models} & \multicolumn{3}{c|}{CNN-based models} & \multicolumn{3}{c|}{Trans-based models} & \multicolumn{3}{c|}{CLIP-based models} & \multicolumn{3}{c|}{IAA-adapted MLLMs} & \multicolumn{4}{c|}{IAA-unadapted MLLMs} & \multicolumn{3}{c}{Ours} \\
          & \footnotesize{TANet} & \footnotesize{ResNext} & \footnotesize{POC}   & \footnotesize{DAT}   & \footnotesize{MaxViT} & \footnotesize{EAT}   & \footnotesize{Comm}   & \footnotesize{AesCLIP} & \footnotesize{CSKD}  & \footnotesize{VILA}  & \footnotesize{Q-Align} & \footnotesize{UNIAA} & \footnotesize{LLaVA-1.5} & \footnotesize{BLIP2} & \footnotesize{miniGPT4} & \footnotesize{GPT-4v} & \footnotesize{CALM}  & \footnotesize{CALM-E} & \footnotesize{CALM-E} \\
    \midrule
    Reso. & 224   & 512   & 640   & 224   & 512   & 224   & 224   & 224   & 224   & 224   & 448   & 336   & 336   & 224   & 224   & -     & 224   & 224   & 336 \\
    Params & 40M   & 43M   & 1.9B  & 87M   & 31M   & 87M   &   -   &   -   &   -   & 383M  & 8.2B  & 6.9B  & 6.9B  & 2.6B  & 7.5B  & -     & 7.1B  & 7.1B  & 7.1B \\
    FLOPs & -     & -     & -     & 240G  & 120G  & 140G  &   -   &   -   &   -   & -     & -     & 359G  & 359G  & 125G  & 550G  & -     & 770G  & 770G  & 878G \\
    \midrule
    PLCC$\uparrow$  & 0.765 & 0.781 & 0.795 & 0.739 & 0.745 & 0.770 & 0.740 & 0.779 & 0.779 & 0.774 & \cellcolor{darkblue!30} 0.823 & \cellcolor{darkblue!30} 0.838 & 0.083 & 0.145 & 0.087 & 0.412 & \textbf{0.829} & \cellcolor{darkblue!30} 0.836 & \cellcolor{darkblue!30} \textbf{0.852} \\
    SRCC$\uparrow$  & 0.758 & 0.780 & 0.794 & 0.738 & 0.708 & 0.759 & 0.734 & 0.771 & 0.770 & 0.774 & \cellcolor{darkblue!30} 0.819 & \cellcolor{darkblue!30} 0.840 & 0.077 & 0.141 & 0.086 & 0.406 & \textbf{0.815} & \cellcolor{darkblue!30} 0.823 & \cellcolor{darkblue!30} \textbf{0.841} \\
    \bottomrule
    \end{tabular}%
  \label{tab:addlabel}%
  }
\end{table*}%

\begin{table}[t]
  \caption{The comparison results on the AC task.}
  \label{tab:avacaption}
  \centering
  \belowrulesep=0pt
  \aboverulesep=0pt
  \setlength{\tabcolsep}{2pt}
  \resizebox{\columnwidth}{!}{
  \begin{tabular}{c|c|ccccccc}
    \toprule
    Models & Reso.& BLEU-1& BLEU-2& BLEU-3& BLEU-4& ROUGE& CIDEr&METEOR   \\
    \midrule
    LLaVA-1.5&336&0.130&0.058&0.022&0.008&0.123&0.000&0.095 \\
    BLIP2&224&0.205&0.090&0.035&0.013&0.137&0.037&0.045 \\
    miniGPT4&224&0.151&0.066&0.024&0.008&0.077&0.000&0.081 \\
    GPT-4v&-&0.116&0.053&0.021&0.008&0.110&0.000&0.100 \\
    \midrule
    CWS&-&0.535&0.282&0.150&0.074&0.254&0.059&0.107\\
    Yeo&-&0.464&0.238&0.122&0.063&0.262&0.051&-\\
    VILA&224&0.503&0.288&0.170&0.113&0.262&0.076&-  \\
    \bottomrule
    CALM &224 & \textbf{0.556}&\textbf{0.335}&{\bf 0.196}&{\bf 0.114}&{\bf 0.286}&{\bf 0.124} & {\bf 0.135}  \\
    CALM-E &224 & \textbf{0.577}&\textbf{0.348}&{0.204}&{0.121}&{0.289}&{0.160} & {0.130}  \\
    CALM-E &336 & {0.558}&{0.345}&\textbf{0.211}&\textbf{0.132}&\textbf{0.295}&\textbf{0.167} & \textbf{0.140}  \\
  \bottomrule
  \end{tabular}
  }
\end{table}

\textbf{Evaluation Metrics.} For AS, we use Spearman Rank-order Correlation Coefficient (SRCC) and Pearson Linear Correlation Coefficient (PLCC) as metrics. SRCC and PLCC measure the ranking accuracy and linear correlation between the predictions and the ground truth, respectively. For AC, we employ BLEU, ROUGE, CIDEr, and METEOR. BLEU and ROUGE focus on the precision of generated words. CIDEr underscores semantic alignment. METEOR accounts for both semantic and structural similarity. For PIAA, SRCC is adopted again as the primary metric.

\textbf{Aesthetic Data Extension.} \textbf{CALM} strictly followed the procedure outlined in Sec. \ref{subsec:comprehensive} for a fair comparison. Besides, we enhanced CALM by using a number of generic and aesthetic question-answer (QA) data during the aesthetic commenting fine-tuning, resulting in an extended version named \textbf{CALM-E}. The generic QA dataset includes CC \cite{lin2014microsoft}, GQA \cite{hudson2019gqa}, OCR-VQA \cite{mishra2019ocr}, TextVQA \cite{singh2019towards}, and VG \cite{krishna2017visual}, while the aesthetic QA dataset is constructed based on their inherent labels through the process described in the \textit{Appendix B}.

\subsection{Comparison to Alternative Approaches}

\textbf{Aesthetic Scoring.} Tab. \ref{tab:avascore} presents a comparison with recent benchmark methods on the AVA dataset. The CNN-based models contain TANet, ResNext and POC \cite{hou2022distilling}, which have been effective so far. The transformer-based models include DAT, MaxViT and EAT, which resort to ViT for IAA. The CLIP-based models consist of Comm, AesCLIP, and CSKD \cite{xu2023clip}, which benefit from language-image pairs. The MLLM-based models are categorized into IAA-adapted MLLMs and IAA-unadapted MLLMs according to whether they are fine-tuned on IAA data. The IAA-adapted MLLMs are composed of VILA, Q-Align, and UNIAA. For the IAA-unadapted MLLMs, we deployed BLIP2, LLaVA-1.5, and miniGPT4 \cite{zhu2023minigpt} locally and called GPT-4v remotely. Because GPT-4v often refuses to answer subjective questions, we only evaluate 7992 images for AS and 5131 images for AC.

When using the same aesthetic labeled data as others, CALM achieves a PLCC of 0.829 and an SRCC of 0.815. Compared to VILA, which is the best IAA-adapted MLLM, CALM shows improvements of 0.055 in PLCC and 0.041 in SRCC. Furthermore, despite evidence from EAT indicating that the higher resolution yields better results, CALM performs better at a smaller resolution than POC, suggesting room for improvement with higher resolution. Additionally, after introducing extra constructed data, similar to Q-Align and UNIAA, and using 336x336 images, CALM-E achieves enhancements of 0.023 in PLCC and 0.026 in SRCC. In conclusion, our CALM-E outperforms all related works and sets a new benchmark in the AS task.

\textbf{Aesthetic Commenting.} Tab. \ref{tab:avacaption} presents a comparative analysis of our method and previous approaches on the AC task. CWS and Yeo integrate CNN features with an LSTM to generate captions, while VILA relies on CoCa to accomplish it. Experimental results indicate that CALM surpasses all methods in both word prediction and semantic alignment. As expected, the inclusion of constructed QA data and the increase in input resolutions enhance the model's capacity for aesthetic discernment and linguistic articulation.

\begin{table}[t]
  \centering
  \belowrulesep=0pt
  \aboverulesep=0pt
  \caption{The comparison results on the PIAA task.}
  \resizebox{\columnwidth}{!}{
    \begin{tabular}{cc|cc}
    \toprule
    \makecell{Models} & SRCC$\uparrow$ & \makecell{Models} & SRCC$\uparrow$ \\
    \midrule
    \makecell{PAM  \cite{ren2017personalized}}&0.520 & \makecell{Wang  \cite{wang2019meta}}&0.522 \\
    \makecell{PA  \cite{li2020personality}}&0.543 & \makecell{BLG  \cite{zhu2020personalized}}&0.561 \\
    \makecell{UG  \cite{lv2021user}}&0.559 & \makecell{SOA  \cite{zhu2021learning}}&0.618 \\
    \makecell{TAPP  \cite{li2022transductive}}&0.591 & \makecell{Hou  \cite{hou2022interaction}}&0.620 \\
    \makecell{MIR  \cite{zhu2022personalized}}&0.621 & \makecell{AFF  \cite{zhu2023personalized}}&0.628 \\
    \midrule
    \makecell{CALM}&0.632 & CALM-In&0.612 \\
    \bottomrule
    \end{tabular}%
  }
  \label{tab:piaa}%
\end{table}%

\textbf{Personalized Image Aesthetic Assessment.} Previous PIAA methods often design an additional network to learn user preferences and guide the IAA backbone to produce personalized scores. In contrast, CALM accomplishes this without any additional network. Notably, we are the first to introduce MLLMs to PIAA. Following the official protocol, we include 10 images per annotator from the test set into the training set and reserve the remaining images for testing. Subsequently, 7 images are used to construct image-score pairs to optimize the prediction of the remaining 3 images. Tab. \ref{tab:piaa} displays the test outcomes, with CALM modestly surpassing the current best model (CALM 0.632 vs. AFF 0.628). Besides, the in-context learning version of CALM, i.e., CALM-In, achieves results on par with the leading models, which will be explored in depth in the ablation study.

\begin{figure*}[t]
    \centering
    \includegraphics[width=\textwidth]{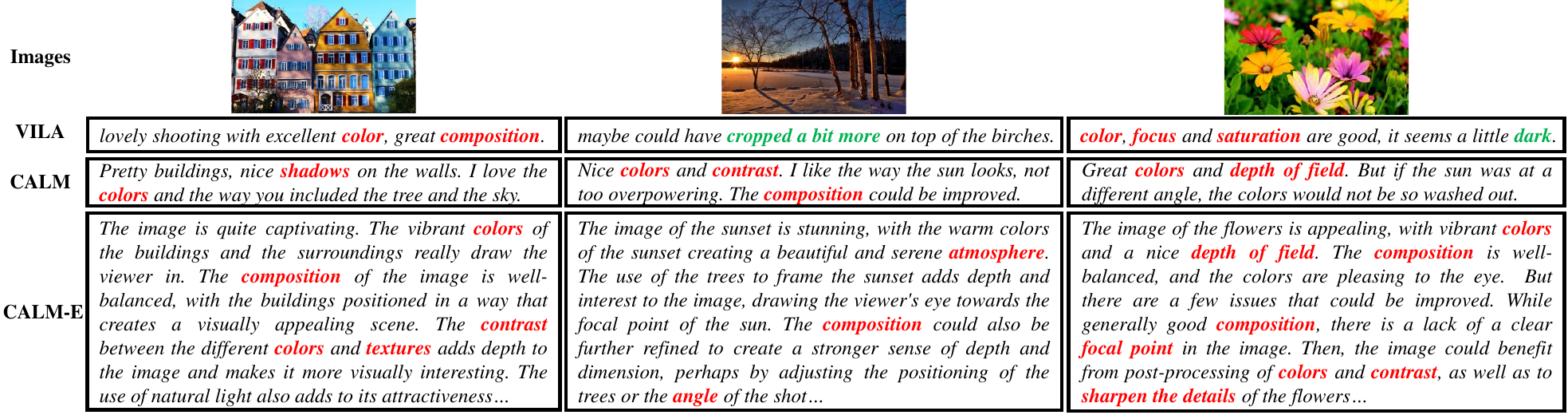}
    \caption{Qualitative comparison of aesthetic commenting. The \textcolor{red}{red} comments are correct, while the \textcolor{green}{green} ones are wrong.}
    \label{fig:vis_comment}
\end{figure*}

\textbf{Qualitative Results.} Fig. \ref{fig:vis_comment} depicts the aesthetic comments produced by VILA, CALM, and CALM-E. VILA offers basic remarks and omits a deeper analysis. CALM provides more elaborate descriptions but misses finer details. In contrast, CALM-E delivers a detailed analysis and actionable suggestions, showcasing a profound grasp of aesthetic principles. Additionally, more MLLMs are evaluated in the \textit{Appendix C}, which includes GPT-4v, qwen-vl \cite{bai2023qwen}, spark-multi-3 \cite{spark2024kedaxunfei}, cogvlm \cite{wang2023cogvlm}, and glm-4v \cite{du2021glm}.

\subsection{Exploration of Zero-shot Aesthetic Suggesting}

To assess zero-shot aesthetic analysis capabilities, we are the first to define the challenging task of aesthetic suggesting. This task requires models to provide suggestions for enhancing the aesthetic value of input images. However, due to the numerous factors that influence aesthetics, evaluating the quality of generated suggestions is challenging. Therefore, we artificially impose a severe degradation on each image to prompt the model to suggest improvements for this degradation. Naturally, the model is permitted to include additional suggestions, as these may also impact the overall aesthetics.

For testing, we curated 100 high-quality images from the PCCD dataset and subjected them to six drastic degradations. These degradations include adding Gaussian or salt-and-pepper noise, applying motion or defocus blur, reducing brightness, increasing brightness, reducing color saturation, and cropping the image. We utilize the instructions shown in Fig. \ref{fig:instruct} to query the MLLMs, and evaluate their responses via GPT-3.5. Specifically, because VILA can only provide a few simple words, we assess its correctness based on whether these words contain the expected degraded attributes.

\begin{table}[t]
  \centering
  \belowrulesep=0pt
  \aboverulesep=0pt
  \setlength{\tabcolsep}{2pt}
  \caption{Comparative accuracy of aesthetic suggesting.}
  \resizebox{\columnwidth}{!}{
    \begin{tabular}{c|cccccc|c}
    \toprule
    \makecell{Types of\\Degradations}  & \makecell{Gaussian or\\Salt-and-\\pepper Noise} & \makecell{Motion or\\Defocus \\ Blur} & \makecell{Brightness \\ Reduction} & \makecell{Brightness \\ Increase} & \makecell{Color\\Saturation \\Reduction} & Cropping & Avg. \\
    \midrule
    VILA & 0.18 & 0.42 & 0.48 & 0.38 & 0.17 & 0.88 & 0.42 \\
    LLaVA-1.5 & 0.42 & 0.54 & 0.29 & 0.07 & 0.47 & \textbf{1.00} & 0.47 \\
    GPT-4v & 0.63 & 0.75 & 0.65 & 0.43 & 0.32 & 0.80 & 0.60 \\
    qwen-vl & 0.08 & 0.43 & 0.22 & 0.09 & 0.16 & 0.84 & 0.30 \\
    spark-multi-3 & 0.07 & 0.39 & 0.32 & 0.09 & 0.21 & 0.98 & 0.34 \\
    cogvlm & 0.12 & 0.44 & 0.26 & 0.04 & 0.25 & 0.89 & 0.33 \\
    glm-4v & 0.01 & 0.26 & 0.22 & 0.00 & 0.17 & 0.98 & 0.27 \\
    \midrule
    CALM      & 0.40 & 0.44 & 0.69 & 0.37 & 0.18 & 0.91 & 0.50 \\
    CALM-E    & \textbf{0.76} & \textbf{0.82} & \textbf{0.89} & \textbf{0.63} & \textbf{0.64} & \textbf{1.00} & \textbf{0.79} \\
    \bottomrule
    \end{tabular}%
  }
  \label{tab:suggestion}%
\end{table}%

\textbf{Quantitative Results.} Tab. \ref{tab:suggestion} displays the accuracy of aesthetic suggesting and shows CALM-E achieves the highest accuracy. Trailing closely, GPT4v and CALM exhibit comparable accuracy. While other MLLMs may provide good suggestions for high-level degradations, they fall short in offering advice on image quality and color. Besides, CALM-E significantly outperforms CALM, confirming that ample generic and aesthetic data can enhance aesthetic insight.

\begin{figure*}[t]
    \centering
    \includegraphics[width=1.0\textwidth]{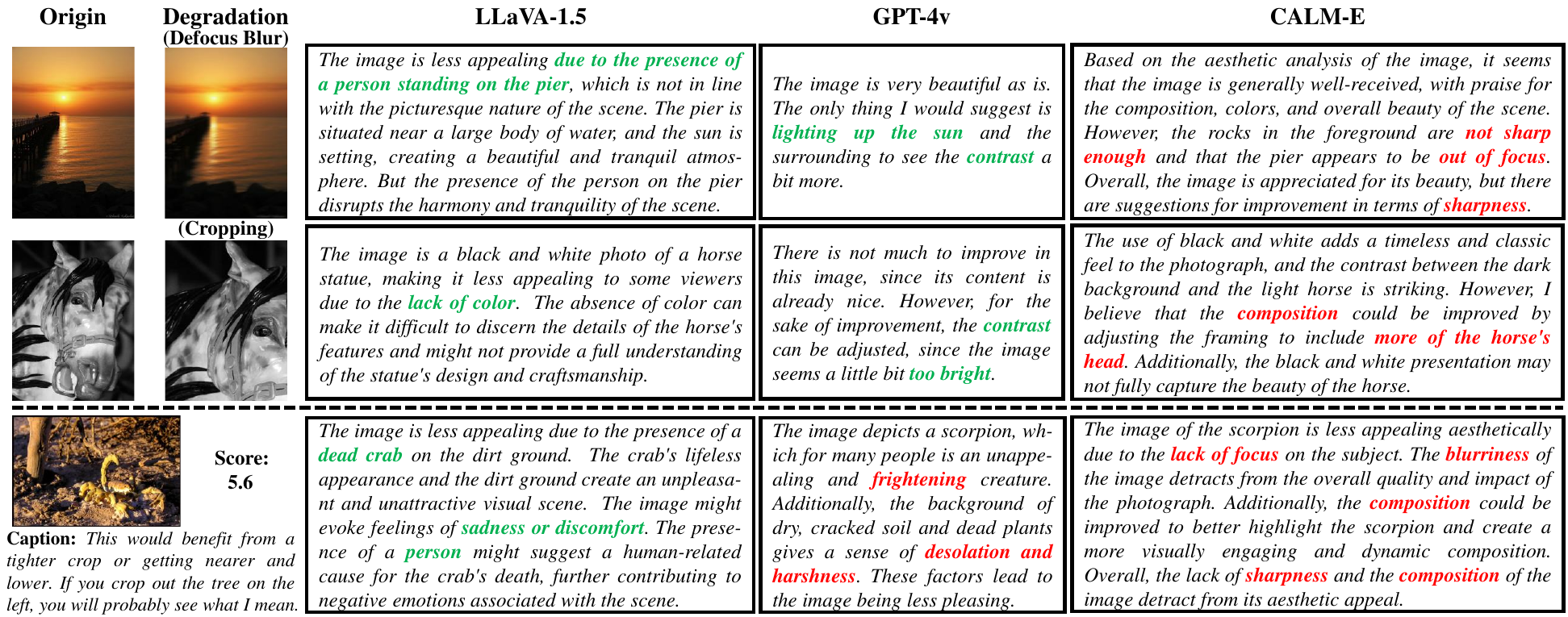}
    \caption{Qualitative comparison of aesthetic suggesting. The \textcolor{red}{red} comments are correct, while the \textcolor{green}{green} ones are wrong.}
    \label{fig:vis_suggestions}
\end{figure*}

\textbf{Qualitative Results.} Fig. \ref{fig:vis_suggestions} presents a comparison of generated suggestions for two degraded images and a real image from the AVA-Captions dataset. Since the real image lacks definitive answers, we include its annotated score and comment for reference. The outcomes reveal that LLaVA-1.5 occasionally hallucinates, and GPT-4v often misses the point. In contrast, CALM-E accurately identifies issues and articulates precise suggestions. Furthermore, we evaluate our approach against additional MLLMs regarding aesthetic suggesting in the \textit{Appendix D}.

\subsection{Ablation Study}

\begin{table}[t]
  \centering
  \belowrulesep=0pt
  \aboverulesep=0pt
  \setlength{\tabcolsep}{3pt}
  \caption{Effects of Qformers at different levels in MFAM.}
  \resizebox{\columnwidth}{!}{
    \begin{tabular}{cccc|cc}
    \toprule
    thematic-level & high-level & middle-level & low-level & PLCC$\uparrow$  & SRCC$\uparrow$ \\
    \midrule
    \checkmark      &       &       &       & 0.768 & 0.759 \\
    \checkmark      & \checkmark      & \checkmark      &       & 0.786 & 0.778 \\
    \checkmark      &       & \checkmark      & \checkmark      & 0.786 & 0.773 \\
    \checkmark      & \checkmark      & \checkmark      & \checkmark      & \textbf{0.829} & \textbf{0.815} \\
    \bottomrule
    \end{tabular}%
  }
  \label{tab:compare_multiscale}%
\end{table}%

\textbf{Does the MFAM help?} To investigate this, we conducted trials by maintaining different Qformers within the MFAM and compared their effects on the AS task. Tab. \ref{tab:compare_multiscale} presents the comparative results. Our baseline maintained only the thematic-level projection, closely resembling LLaVA-1.5. However, with the addition of each layer of Qformers, a notable improvement was observed--—a boost by 0.061 in PLCC and 0.056 in SRCC. This clearly demonstrates the necessity of aligning features across all three levels.

\textbf{How many aesthetic queries in Qformers are optimal?} Intuitively, the more the aesthetic queries, the higher the accuracy and computational cost will be. To explore the trade-off between effectiveness and performance, we conducted an ablation study on the AC task. As shown in Fig. \ref{tab:compare_number_querytoken}, with \#queries increasing, the effect improves rapidly at first and then stabilizes when the number reaches 32. Besides, due to the attention operations in Qformers and the LLM, the overall computational cost increases quadratically, becoming noticeable if \#queries is large. Therefore, we opted for 32 queries per Qformer. Of course, having more queries per Qformer may be better if computing resources are sufficient.

\textbf{Is the text-guided self-supervised learning useful?} To answer this, we removed the self-supervised pre-training from the standard training process for AS task. Evaluation results are shown in the first column of Tab. \ref{tab:compare_selfsupervise}. Compared with the original CALM, CALM without self-supervised learning underperforms by 0.047 in PLCC and by 0.041 in SRCC. Although MFAM structurally facilitates multi-scale feature extraction, the absence of text-guided self-supervised learning significantly impairs the overall effects.

\begin{table}[t]
  \centering
  \belowrulesep=0pt
  \aboverulesep=0pt
  \setlength{\tabcolsep}{3pt}
  \caption{Effects vary with \#queries in Qformer on AC task.}
  \resizebox{\columnwidth}{!}{
    \begin{tabular}{c|ccccccc}
    \toprule
    \#queries & BLEU-1 & BLEU-2 & BLEU-3 & BLEU-4 & ROUGE & CIDEr & METEOR \\
    \midrule
    4 & 0.509 & 0.290 & 0.157 & 0.080 & 0.264 & 0.094 & 0.123 \\
    8 & 0.537 & 0.321 & 0.183 & 0.099 & 0.277 & 0.112 & 0.128 \\
    16 & 0.544 & 0.327 & 0.188 & 0.103 & 0.284 & 0.116 & 0.129 \\
    32 & \textbf{0.556} & 0.335 & 0.196 & 0.114 & 0.286 & 0.124 & \textbf{0.135} \\
    64 & 0.551 & \textbf{0.336} & \textbf{0.199} & \textbf{0.118} & \textbf{0.289} & \textbf{0.142} & 0.132 \\
    \bottomrule
    \end{tabular}%
  }
  \label{tab:compare_number_querytoken}%
\end{table}%

\textbf{Is every type of data augmentation necessary?} To delve deeper into the role of each data augmentation type for self-supervised learning, we categorize them into three groups: quality (noise, compression, blur, pixelation), color (brightness, saturation, contrast), and subject (blurring or masking foreground, cropping objects). We then conducted trials on AS task using various combinations of these augmentations. Tab. \ref{tab:compare_selfsupervise} reveals nuanced insights into the impact of each augmentation type. The findings suggest that solely leveraging quality or color augmentations yields modest improvements, and relying exclusively on subject augmentations appears to provide no significant benefit, likely due to the visual encoder's inherent high-level reasoning capacities. However, a synergy is observed when subject augmentations are introduced alongside quality and color augmentations. This integration not only contributes to improved outcomes but also ensures the preservation of high-level information.

\textbf{How well is in-context learning suited for PIAA?} In-context learning involves providing a model with QA examples of similar questions before asking a specific question, thereby enabling the model to answer such questions. Theoretically, it is ideal for solving the PIAA task, as it allows the model to infer a user's implicit aesthetic tastes from the QA examples. To test this hypothesis, instead of adding 10 images per annotator from the test set to the training set, we randomly select 5 images per annotator to construct the in-context instructions shown in Fig. \ref{fig:instruct} for use during the test period. Note that CALM-In still requires training on the FLICKR-AES training set; otherwise, it does not perform well on the in-context learning-based PIAA task. Tab. \ref{tab:piaa} demonstrates that, despite not acquiring the annotators' preferences in advance, CALM-In can elicit user preferences and achieve outcomes comparable to some latest methods.

\textbf{Are there any limitations?} Firstly, Tab. \ref{tab:avascore} indicates that increasing image resolution can enhance the effect of the AS task. However, our exploration in this aspect is limited, as our visual encoder, the pre-trained ViT-L/14, cannot accommodate varying resolution inputs as flexibly as CNNs. Secondly, the computational burden of our approach is somewhat high, which may limit its applicability scenarios. We intend to address these identified shortcomings in the future.

\begin{table}[t]
  \centering
  \belowrulesep=0pt
  \aboverulesep=0pt
  \caption{Comparison of using different data augmentation types in the text-guided self-supervised learning on AS task.}
  \resizebox{\columnwidth}{!}{
    \begin{tabular}{c|cccccccccc}
    \toprule
    quality & & \checkmark & & & \checkmark & \checkmark & & \checkmark \\
    color   & & & \checkmark & & \checkmark & & \checkmark & \checkmark \\
    subject & & & & \checkmark & & \checkmark & \checkmark & \checkmark \\
    \midrule
    PLCC$\uparrow$ & 0.782 & 0.789 & 0.789 & 0.716 & 0.807 & 0.788 & 0.788 & \textbf{0.829} \\
    SRCC$\uparrow$ & 0.774 & 0.776 & 0.779 & 0.780 & 0.793 & 0.776 & 0.775 & \textbf{0.815} \\
    \bottomrule
    \end{tabular}%
  }
  \label{tab:compare_selfsupervise}%
\end{table}%

\section{Conclusion}

Our study presents CALM, an advanced comprehensive aesthetic large language model. To ensure the extraction of multi-scale aesthetic features both structurally and functionally, we propose the multi-scale text-guided self-supervised learning. Additionally, the instruct-tuning technique is developed to enable CALM to perform multiple aesthetic tasks. Extensive testing reveals that CALM outperforms the current leading approaches across all IAA tasks, solidifying its dominance in the field of IAA. Furthermore, its remarkable zero-shot capabilities in in-context learning PIAA and offering aesthetic suggestions are fully exploited.

\bibliography{aaai25}



\section*{Supplementary material}

# Appendix

## A. Image Augmentations in Text-Guided Self-Supervised Learning

In Table 1, we document the image augmentations, their parameters, and the corresponding guided textual pseudo-labels used during the self-supervised pre-training. To conserve space, we omit the pseudo-labels for more intense augmentations applied to the second image. These augmentations are categorized into three types: degradation of image quality (e.g., Gaussian blur, impulse noise, JPEG compression, pixelate, motion blur, defocus blur), alteration of image color (e.g., brightness, saturation, contrast adjustments), and modification of image content (e.g., subject or non-subject object masking, foreground or background blurring). Additionally, we provide guided textual pseudo-labels for cases where the same augmentations are applied to both images.

## B. Aesthetic QA Data Construction

In this section, illustrated in Listings1, we demonstrate how GPT-3.5 can assist in generating aesthetic QA dialogues, using the PCCD dataset as a case study. We start by creating a prompt template that includes variable fields enclosed in angled brackets. Next, we design an example QA dialogue for GPT-3.5 to emulate in style and content. Finally, we replace the placeholders with image-specific information and input this data into GPT-3.5, resulting in a five-round QA dialogue.

## C. Visualization of Aesthetic Comments

In this section, we evaluate the performance of our CALM and CALM-E models against leading MLLMs in the domain of aesthetic commenting, including GPT-4v, qwen-vl, spark-multi-3, cogvlm, and glm-4v. We use a standard question prompt across all models: "Please comment on the current image aesthetically". For illustrative clarity, Fig. 1 through Fig. 8 show each model's output, with comments relevant to aesthetic analysis highlighted in red for easy comparison.

The figures reveal that while CALM is capable of generating aesthetically relevant keywords, its responses tend to be concise. In contrast, CALM-E provides comprehensive evaluations, detailing both strengths and weaknesses in each image's aesthetics and offering suggestions for improvement. GPT-4v, qwen-vl, spark-multi-3, and cogvlm predominantly focus on content description, significantly overlooking aesthetic aspects. Although their comments are generally accurate content-wise, the lack of aesthetic insight is a notable shortcoming. glm-4v occasionally produces relevant aesthetic commentary but often fails to address all aesthetic attributes; for example, Fig. 4 and Fig. 5 show it addressing aesthetics primarily from a content-based perspective.

## D. Visualization of Aesthetic Suggestions

In this section, we evaluate our CALM and CALM-E models against prominent mainstream MLLMs in terms of the aesthetic suggesting task. The comparative panel of MLLMs includes GPT-4v, qwen-vl, spark-multi-3, cogvlm, and glm-4v. Each model was given the identical query: "Please suggest some aesthetic improvements to this image." We present the results in Fig. 9 through Fig. 16. For images subjected to artificial degradation, we indicate the type of degradation in the top right corner. Additionally, we color-code the suggestions: red for those aligning with the standard answer, blue for those deemed highly reasonable upon manual review, and green for those considered highly unreasonable. For unaltered original images, blue highlights very reasonable suggestions, and green notes particularly unreasonable ones upon manual examination.

These visualizations reveal that CALM provides succinct but targeted improvement suggestions, effectively identifying key areas for aesthetic improvement. Moreover, CALM-E delivers comprehensive and detailed advice for each image, thoroughly addressing various aesthetic attributes. Models such as GPT-4v, qwen-vl, spark-multi-3, cogvlm, and glm-4v occasionally align with standard recommendations and offer additional plausible suggestions. However, they also frequently propose numerous unreasonable or irrelevant suggestions that do not pertain to aesthetic comments.

Table 1: Image augmentations, parameters, and the corresponding guided textual pseudo-labels used during the self-supervised pre-training.

| Augmentations | Parameters | Guided text pseudo-labels if the first image is conducted a more severe augmentation |
|---|---|---|
| Gaussian Blur | 1, 2, 3, 4, 5 | "The first image is more noisy.</s>", "The clarity in the first image is worse.</s>", "There"s more noticeable noise in the first image.</s>", "The first image has a greater amount of noise.</s>", "The first image seems like having a worse quality.</s>", "The quality of the first image appears to be lower.</s>", "The first picture looks like it has a poorer quality.</s>", "The first picture seems more blurry.</s>", "The first picture looks fuzzier.</s>", "There"s a bit more blur in the first image.</s>", "The first image appears to be less clear.</s>", "The second picture is a little clearer.</s>" |
| Impulse Noise | 1, 2, 3, 4, 5 | "The first image is more noisy.</s>", "The clarity in the first image is worse.</s>", "There"s more noticeable noise in the first image.</s>", "The first image has a greater amount of noise, may have been corrupted.</s>", "The first image seems like having a worse quality.</s>", "The quality of the first image appears to be lower, it may have been damaged.</s>", "The first picture looks like it has a poorer quality.</s>", "The first picture seems more blurry.</s>", "The first picture looks fuzzier.</s>", "The first image appears to be less clear.</s>", "The second picture is a little clearer.</s>" |
| JPEG Compression | 1, 2, 3, 4, 5 | "The first image seems like having a worse quality, may have been compressed.</s>", "The quality of the first image appears to be lower, it may have been compressed.</s>", "The clarity in the first image is worse.</s>", "The first picture looks like it has a poorer quality.</s>", "The first picture seems more blurry.</s>", "The first picture looks fuzzier.</s>", "There"s a bit more blur in the first image.</s>", "The first image appears to be less clear.</s>", "The second picture is a little clearer.</s>" |
| Pixelate | 1, 2, 3, 4, 5 | The clarity in the first image is worse.</s>", "The first picture is even less clear, and there"s a noticeable decrease in pixel quality.</s>", "The first image looks more unclear and seems to have a lower pixel count.</s>", "In comparison, the first image appears blurrier and has a lower pixel resolution.</s>" "The first image seems to have deteriorated further, displaying both lower clarity and fewer pixels.</s>", "The first image seems like having a worse quality.</s>", "The quality of the first image appears to be lower.</s>", "The first picture looks like it has a poorer quality.</s>", "The first picture seems more blurry.</s>", "The first picture looks fuzzier.</s>", "There"s a bit more blur in the first image.</s>", "The first image appears to be less clear.</s>", "The second picture is a little clearer.</s>", "The second image is sharper and seems to have higher pixels.</s>", "The second image appears sharper and seems to boast a higher pixel count.</s>", "The sharpness in the second image is noticeable, and it seems to possess a higher pixel resolution.</s>", "Comparatively, the second image looks sharper and suggests a higher pixel density.</s>" |

| Type | Parameters | Prompts |
|---|---|---|
| Motion Blur | 1, 2, 3, 4, 5 | "The first one is more blurry.</s>", "The clarity in the first image is worse.</s>", "The first picture seems more blurry.</s>", "The first picture looks fuzzier.</s>", "There"s a bit more blur in the first image.</s>", "The first image seems to lack sharpness.</s>", "The first image seems to exhibit a greater level of blur.</s>", "The blurriness seems more noticeable in the first image.</s>", "The visual impression suggests that the first image is blurrier.</s>", "The initial image appears to have more pronounced blurriness.</s>", "The first image appears to have more severe motion blur.</s>", "The first image appears to be less clear.</s>", "The second picture is a little clearer.</s>", "The second one is clearer, the first one seems a little out of focus.</s>", "The first image has a higher degree of motion blur compared to the second one.</s>", "The first one looks a little defocus.</s>" |
| Defocus Blur | 1, 2, 3, 4, 5 | "The first one is more blurry.</s>", "The clarity in the first image is worse.</s>", "The first picture seems more blurry.</s>", "The first picture looks fuzzier.</s>", "There"s a bit more blur in the first image.</s>", "The first image seems to lack sharpness.</s>", "The first image seems to lack sharpness, the focus seems off.</s>", "The first image seems to exhibit a greater level of blur.</s>", "The blurriness seems more noticeable in the first image.</s>", "The visual impression suggests that the first image is blurrier.</s>", "The initial image appears to have more pronounced blurriness.</s>", "The first image appears to have more severe defocus blur.</s>", "The first image appears to be less clear.</s>", "The second picture is a little clearer.</s>", "The second one is clearer, the first one seems a little out of focus.</s>", "The first image has a higher degree of defocus blur compared to the second one.</s>", "The first one looks a little defocus.</s>" |
| Brightness | 0.3,0.5, 0.7,1, 1.3,1.5, 1.7,1.9, 2.0,3.0 | "The first image seems brighter.</s>", "The first image have a higher level of brightness.</s>", "The second image looks darker.</s>" |
| Saturation | 0.3,0.5, 0.7,1, 1.3,1.5, 1.7,1.9, 2.0,3.0 | "The first image looks more colorful.</s>", "The first image seems to be more saturated.</s>", "The colors are more vibrant and saturated in the first image.</s>", "The first picture appears to have higher saturation.</s>", "The saturation level is higher in the first image.</s>", "The colors in the second image are less vibrant and appear desaturated.</s>", "The second picture seems to have lower saturation.</s>", "The second picture is grayer.</s>", "In contrast, the second image is more black and white.</s>" |
| Contrast | 0.3,0.5, 0.7,1, 1.3,1.5, 1.7,1.9, 2.0,3.0 | "The first image has higher contrast.</s>", "The first image is more contrasting.</s>" "There is a higher level of color contrast in the first picture.</s>", "The first image has a higher level of contrast between its colors.</s>", "The second picture is grayer.</s>", "The second image is less contrasting in color.</s>", "The second image has a lower level of color contrast.</s>" |
| Masking Object | None | "The main subject of this image is occluded.</s>", "The primary subject in this image is obscured or hidden.</s>", "There is an obstruction affecting the main subject in this image.</s>", "The primary subject appears to be covered or hidden in this image.</s>", "The main subject is not fully visible due to an obstruction in this image.</s>" |
| Masking Non-object | None | "The main subject of this image is not occluded.</s>", "The primary subject in this image is clear and not obscured.</s>", "There is no obstruction affecting the main subject in this image.</s>", "The main focus of this image is visible without any occlusion.</s>", "The central subject in this image is not hidden or covered.</s>", "The primary subject is fully visible without any obstruction in this image.</s>" |

| | | |
|---|---|---|
| Foreground Blur | 0,5, 10,25, 50,100 | "The first image"foreground is more noisy.</s>", "There"s more noticeable noise in the first image"foreground.</s>", "The foreground of the first image is more noisy.</s>", "The clarity in the first image"foreground is worse.</s>", "There"s more noticeable noise in the first image"foreground.</s>", "The foreground of teh first image has a greater amount of noise.</s>", "The first picture"foreground seems more blurry.</s>", "The first picture"foreground looks fuzzier.</s>", "There"s a bit more blur in the first image"foreground.</s>", "The foreground of the first image appears to be less clear.</s>", "The second picture"foreground is a little clearer.</s>" |
| Background Blur | 0,5, 10,25, 50,100 | "The first image"background is more noisy.</s>", "There"s more noticeable noise in the first image"background.</s>", "The background of the first image is more noisy.</s>", "The clarity in the first image"background is worse.</s>", "There"s more noticeable noise in the first image"background.</s>", "The background of teh first image has a greater amount of noise.</s>", "The first picture"background seems more blurry.</s>", "The first picture"background looks fuzzier.</s>", "There"s a bit more blur in the first image"background.</s>", "The background of the first image appears to be less clear.</s>", "The second picture"background is a little clearer.</s>" |
| Guided text pseudo-labels if the two images are conducted the same augmentation | | |
| | | "These are two identical images.</s>", "These two pictures are identical.</s>", "The images are the same.</s>", "Both pictures are exactly alike.</s>", "There is no difference between these two pictures.</s>", "The two pictures are identical in every aspect.</s>" |

Listing 1: Aesthetic QA Data Construction

```
## A template is defined for asking GPT-3.5
org_prompt = '''You are an AI visual assistant, and you are seeing a single photography. You are provided with attributes of this photography, including a category, a description and an overall aesthetic score. Additionally, comments on various aspects of the photography are provided, including general impression, composition, depth of field, focus, subject of the photo, use of the camera, color, and lighting. Design a conversation between you and a person asking about this photo. The answers should be in a tone that an AI visual assistant is seeing the image and answering the question. Ask 5 diverse questions and give the corresponding answers. The questions should be related to the aesthetics of the photo. Only include questions that have definite answers.

Example:
<example>

Now respnose the following:

Category: <category>
Description: <description>
Score: <score>
General impression: <general_impression>
Composition: <composition>
Depth of field: <depth_of_field>
Focus: <focus>
Subject of photo: <subject_of_photo>
Use of camera: <use_of_camera>
Color and lighting: <color_lighting>

Conversation:
'''

## An example is provided for GPT-3.5 to follow in generating.
example = '''Category: nature-photography
Description: For the beauty,the morning light
Score: 10
General impression: Hi Angela,Less is more, and you did a nice job of capturing the scene, and the feeling of the place.I want to walk right into the picture! Very inviting. I want to know more, and so you have successfully enticed me, the reader.
```

```
28  Composition: Very nice composition. I like the way the road curves and leads your eye into the picture, a nice 'S'
        curve. I also like that the picture has a front, a middle and a back, and sense of depth. A very clean
        composition. Simple. A very nice relaxing feel.
29  Depth of field: Just great!
30  Focus: The focus is great. Your subject is very sharp, and the background a bit out of focus, which is perfect.
31  Subject of photo: Really neat looking old building, Angela!I wish in your description you had told me a big more—
        what country this was taken in, and what these buildings were used for, or still being used for? And I with
        that information, I would probably call it a travel photo.One of our goals with photography is tell a story, or
        share an experience, and so providing some background is important.
32  Use of camera: You did a great job of framing the picture, and the exposure looks great. It looks like you might have
         played with the contrast a bit, or maybe done some HDR? I say that because there is more detail in the shadows
        than I would usually see. I think it works well in the main part of the picture. In the hills behind, if you
        opened up those shadows, I don't like it as well there. I would like to see the hills go darker—and let them
        fade into the background and just be a shape in the distance.
33  Color and lighting: Beautiful colors, nice muted, relaxing palette. The light is nice. If this scene is close to you,
         I would love to see you go back and shoot it several times in early morning light to see if you get an even
        more dramatic sunrise.
34
35  Conversation:
36  Question 1:
37  User: "Can you tell me about the general impression of this nature photography?"
38  AI: "Certainly! The photo exudes a serene and inviting atmosphere, capturing the essence of the scene effectively.
        Its simplicity adds to its charm, making one feel like they want to step into the picture and explore the area.
        It's an enticing image that piques curiosity about the location."
39
40  Question 2:
41  User: "How would you describe the composition of this image?"
42  AI: "The composition is notably well-executed. It features a pleasing 'S' curve created by the road, leading the
        viewer's eye into the scene. The photo is structured into three distinct segments — a foreground, a middle
        ground, and a background, providing a clear sense of depth. This clean and simple composition contributes to the
         photograph's relaxing feel."
43
44  Question 3:
45  User: "What about the depth of field in this photo?"
46  AI: "The depth of field is expertly managed. It enhances the sharpness of the subject while softly blurring the
        background. This technique ensures that the main subject stands out clearly, drawing attention to the key
        elements of the photo without any distractions."
47
48  Question 4:
49  User: "How effective is the use of color and lighting in this photo?"
50  AI: "The color palette is beautifully muted, creating a tranquil and relaxing visual experience. The lighting,
        possibly captured in the morning, adds a soft, natural glow to the scene. The suggestion to revisit the location
         for different sunrise effects is a great idea to explore varying lighting conditions for an even more dramatic
        impact."
51
52  Question 5:
53  User: "Can you comment on the use of the camera in this photograph?"
54  AI: "The photographer has skillfully framed the picture, and the exposure is well-balanced. There are indications of
        post-processing, such as enhanced contrast or HDR techniques, which bring out details in the shadows. This
        approach works well in the main part of the image, though the hills in the background could benefit from a
        darker tone to fade into the distance and add a sense of mystery."'''
55
56  ## Get the information of a photo from the PCCD dataset
57  descirption = item['description']
58  score = item['overall']
59  category = item['category']
60  general_impression = item['general_impression']
61  composition = item['composition']
62  depth_of_field = item['depth_of_field']
63  focus = item['focus']
64  subject_of_photo = item['subject_of_photo']
65  use_of_camera = item['use_of_camera']
66  color_lighting = item['color_lighting']
67
68  ## make the final prompt to ask GPT-3.5
```

```
69  prompt = org_prompt.replace('<category>',category).replace('<description>',desicrption).replace('<score>',score).
            replace('<general_impression>',general_impression).replace('<composition>',composition).replace('<depth_of_field
            >',depth_of_field).replace('<focus>',focus).replace('<subject_of_photo>',subject_of_photo).replace('<
            use_of_camera>',use_of_camera).replace('<color_lighting>',color_lighting).replace('<example>',example))
70  response = get_response(prompt)
```

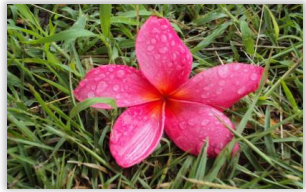

| | |
|---|---|
| **CALM** | *I **like** the color and the water droplets, but I think the grass blades in the background are a little **distracting**. Nobody likes an "out of focus" subject in a shot like this.* |
| **CALM-E** | *The image of the flower is visually **appealing**, with a vibrant and eye-catching color palette. The contrast between the vivid pink of the flowers and the green of the grass adds a vibrant feel to the image. The water droplets on the petals add a dynamic element to the image, capturing the delicate beauty of the flower. However, the background of the image is somewhat **distracting**, with a busy and cluttered appearance. The presence of the grass in the background detracts from the overall aesthetic of the image, drawing attention away from the main subject. **A more minimalist or blurred background would have allowed the flower to stand out more prominently**, enhancing the overall visual impact of the image. Additionally, the focus of the image appears to be slightly off, with the flower and water droplets not being as sharply in focus as they could be. This lack of sharpness detracts from the overall clarity and detail of the image, diminishing its overall aesthetic appeal. Overall, while the image of the flower is visually striking and captures the beauty of nature, the distracting background and lack of sharp focus detract from its overall aesthetic appeal. With some adjustments to the background and focus, the image could be even more visually impactful and aesthetically pleasing.* |
| **GPT-4v** | *The deep pink color of the flower contrasts **beautifully** with the green grass in the background. The water droplets on the petals add a fresh and natural element to the image, making it feel as though the flower has just been kissed by rain.* |
| **qwen-vl** | *The image features a single pink flower lying on a bed of green grass. The flower has a red central ring and pink petals, with small red stamens in the center. It appears to be a frangipani flower. The flower is gently curving on the grass, which has a few water droplets on it. From a visual perspective, the combination of the vibrant pink color of the flower and the green grass creates a **pleasing contrast and a sense of freshness**. The addition of water droplets on the grass further enhances the natural and refreshing feel of the scene.* |
| **spark-multi-3** | *This image shows a pink flower with water droplets on its petals. The flowers look very **beautiful** in the background of the green grass.* |
| **cogvlm** | *The picture shows a pink flower lying on the grass. Specifically, the dew is on the petals of a pink flower, and the grass around it is green.* |
| **glm-4v** | *This photo is aesthetically **pleasing** because it captures the beauty of nature. The vibrant pink color of the flower contrasts beautifully with the green grass, creating a visually stunning scene. Additionally, water droplets on the petals add an element of freshness and delicacy to the flower, making it even more attractive.* |

Figure 1: The first comparison of aesthetic comments.

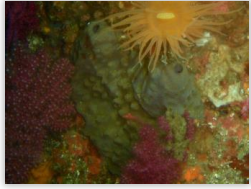

| | |
|---|---|
| **CALM** | *It's very hard to take photos in an aquarium, I've tried, so I know ;-) You've done a better job than I have, but the **colours are a bit dull** and **the contrast is almost non existant**. Nobody has said that they like the photo, so I guess you can't please everyone ;-)* |
| **CALM-E** | *The image subject of the sea creature is **appealing**. Colorful Marine life makes for an intriguing scene. However, the image does suffer from some technical issues. **The focus is not sharp**, and the overall image appears to be **out of focus**. This detracts from the overall impact of the image and makes it difficult to fully appreciate the details of the sea creature. Additionally, the image appears to be quite **noisy**, which further diminishes the overall quality of the photograph. Moreover, the lighting of the picture appears to be somewhat **dim**, adjusting the lighting can make the color contrast of Marine life more prominent and more attractive. In terms of aesthetics, the image could benefit from a sharper focus and a reduction in noise. This would help to bring out the details of the sea creature and create a more visually appealing image. Additionally, **adjusting the composition to create a more dynamic and engaging visual narrative** could enhance the overall impact of the photograph. Overall, while the image has some visually appealing elements, the technical issues detract from its overall impact. With some adjustments to the focus, noise, and composition, the image has the potential to be a visually stunning and captivating photograph.* |
| **GPT-4v** | *This underwater image captures a **beautiful** sea anemone with its tentacles extended, surrounded by colorful marine life. The vibrant shades of pink, purple, and orange create a **mesmerizing** scene, showcasing the diversity and beauty of the ocean's ecosystem. The texture of the anemone and the surrounding coral and sea sponges add depth to the image, inviting the viewer to explore the intricate details of this underwater landscape.* |
| **qwen-vl** | *The image features a **beautiful** underwater sea scene with a variety of colorful sea creatures, including a large sea anemone in bright orange with white polyps, a green sea slug, a purple and orange sea coral, and a yellow sea sponge all thriving in the rich sea habitat. The different shapes, colors, and textures of the sea creatures create a striking and visually appealing contrast against the dark background of the sea floor. This image showcases the natural beauty and biodiversity of the ocean ecosystem.* |
| **spark-multi-3** | *This image shows an ocean landscape that includes a variety of corals and anemones of different colors. These creatures exhibit vivid colors such as purple, yellow, orange, and red, among others. They grow freely in the seawater and form a beautiful and unique ecosystem. The whole scene gives a **quiet and mysterious feeling**, as if you are in a magical underwater world.* |
| **cogvlm** | *The picture shows a **colorful** underwater scene. In the middle of the picture is a large yellow flower with many tentacles, and to its left is a gray rock with two holes. Underneath the rock are pink and purple corals.* |
| **glm-4v** | *This picture is **beautiful** because it shows a rich and diverse marine ecosystem. The colorful coral reefs and sea anemones create a stunning visual effect, making people feel immersed in the ocean. In addition, this photo can also **inspire people to explore the mysteries** of the natural world and appreciate the beauty of life in the sea.* |

Figure 2: The second comparison of aesthetic comments.

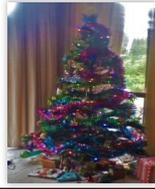

| | |
|---|---|
| **CALM** | *I would have liked to see the tree in **focus**, with the rest of the room slightly **blurred**.* |
| **CALM-E** | *The image of the holiday decorations is visually **appealing**, with a festive and cheerful atmosphere. The use of bright and vibrant colors adds to the overall aesthetic appeal of the image. However, the image appears to be **blurry**, which detracts from its overall quality. The lack of sharpness and clarity in the image makes it difficult to fully appreciate the details of the decorations. The image seems to have been taken with a **low-quality camera**, as evidenced by the grainy and pixelated appearance. There are also harsh spots at the top of the image caused by overexposure. Additionally, the Christmas tree is slightly **skewed** and the decorations underneath are **distracting**. To improve the aesthetic quality of the image, it would be beneficial to use a **higher quality camera or to ensure that the image is captured with better focus and clarity**. Additionally, **adjusting the lighting and contrast levels** could help to enhance the overall visual appeal of the decorations. Overall, while the image captures the festive spirit of the holiday season, the lack of sharpness and clarity detracts from its aesthetic appeal. With some adjustments to the technical aspects of the image, it has the potential to be a visually stunning representation of holiday decorations.* |
| **GPT-4v** | *The image depicts a beautifully decorated Christmas tree with multicolored lights and shiny tinsel. There are various colorful ornaments hung on the branches, adding to the **festive feel**. Underneath the tree, there are several wrapped presents, indicating that this may be a family home during the holiday season. The natural light coming in from the window behind the tree gives the room a warm and cozy atmosphere. Overall, the image captures the **essence of Christmas joy and the excitement of gift-giving**.* |
| **qwen-vl** | *The image features a lush Christmas tree adorned with colorful lights and tinsel, standing in a room near a window. The tree is wrapped in a multi-colored garland, making it a **vibrant** centerpiece of the scene. The colorful lights add warmth and a festive touch to the room, creating a cozy atmosphere for the holiday season. The tinsel further enhances the tree's visual appeal, making it more eye-catching and cheerful. The tree is also surrounded by numerous presents, indicating that it is a special occasion, and the gifts signify the spirit of giving and celebration. Overall, the image portrays a beautifully decorated Christmas tree that captures the **essence of the holiday season**.* |
| **spark-multi-3** | *The picture shows a beautifully decorated Christmas tree with colorful decorations and lights. Around the tree, we can see some gift boxes scattered on the ground, some of which have been opened. The whole scene creates a **warm and cheerful atmosphere**, which makes people feel the joy and warmth of Christmas.* |
| **cogvlm** | *A Christmas tree stands in the corner of a room. There are many colorful lights on the tree, and several packages under the tree. In the background is a window with yellow curtains.* |
| **glm-4v** | *This photo is **not aesthetically pleasing** because the Christmas tree looks **messy** and the colors are too **bright**.* |

Figure 3: The third comparison of aesthetic comments.

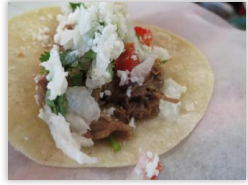

| | |
|---|---|
| **CALM** | *I'm sorry, but this looks **unappetizing**. It looks like you've got a bit of a **color cast** on as well. I'm afraid this looks like it was taken with a very low-end camera phone.* |
| **CALM-E** | *This image presents a close-up of a taco. Due to several factors, the image is **less appealing** aesthetically. Firstly, **the lack of sharpness** in the image detracts from its overall visual appeal. The focus of the image appears to be slightly soft, which detracts from the overall sharpness and clarity of the taco. The soft focus makes it difficult to discern the details of the food, which is essential for a recipe food image. Additionally, the **composition** of the image could be improved. It would have been **more effective if a tighter crop was applied**. The current composition feels a bit off, and a tighter crop would have helped to draw more attention to the food itself. Furthermore, **the lack of contrast and light** in the image is also mentioned as a drawback. A little more contrast would have helped to make the food stand out more. Adding a warm light source or adjusting the saturation during post-processing may increase the overall visual appeal. Overall, while the food itself looks delicious, the image could benefit from improvements in sharpness, composition, and contrast to make it more visually appealing.* |
| **GPT-4v** | *This is a close-up image of a delicious looking taco with a corn tortilla filled with tender shredded meat, topped with crumbled white cheese, chopped cilantro, and diced tomatoes. The ingredients look **fresh and flavorful**, making it a tempting choice for a quick and satisfying meal.* |
| **qwen-vl** | *The image features a close-up view of a small corn tortilla topped with a delicious-looking beef taco. The tortilla, which is the main component of the taco, is a bright yellow color. The filling, which includes pulled beef, tomatoes, cheese, and onions, adds a burst of color to the dish, with the red tomatoes and green onions standing out prominently. The cheese, which is likely feta or queso fresco, adds a creamy texture to the taco, making it visually appealing. Overall, the image presents a **well-crafted and appetizing** taco that showcases the combination of colors and textures that make it a delicious and inviting meal.* |
| **spark-multi-3** | *This image shows a delicious burrito topped with ground meat, lettuce, tomato, and sour cream. The whole burrito looks very attractive, colorful and layered. The green lettuce leaves contrasted sharply with the red tomatoes. Ground meat and sour cream add richness and flavor to the burrito. Overall, this is a **mouth-watering** food image.* |
| **cogvlm** | *A delicious taco is placed on a white plate. The taco has a soft shell, and the filling consists of shredded meat, diced tomatoes, onions, cilantro, and a sprinkle of cheese.* |
| **glm-4v** | *This photo is aesthetically **pleasing** because it shows a **delicious** taco that looks appetizing. The colors are vibrant and the ingredients look fresh. The presentation is also attractive, with the taco placed on white paper, emphasizing its colors and textures. Overall, the image evokes a sense of appetite and makes one want to try the taco described in the article.* |

Figure 4: The fourth comparison of aesthetic comments.

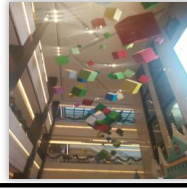

| | |
|---|---|
| **CALM** | *I think this has a lot of **potential**, but the glare on the left is very **distracting**. Nobody is going to look past that. Also, it looks a little soft, like it wasn't sharp enough in the camera?* |
| **CALM-E** | *This image shows the interior of a shopping mall. The aesthetic quality of this image is primarily contributed to by the **vibrant and colorful** display of the square. The contrast between the different colors adds depth and dimension to the image, making it visually stimulating. However, the image is less appealing due to several factors. The first issue is the **lack of focus**, which makes it difficult for the viewer to discern the subject of the image. This lack of clarity detracts from the overall visual impact of the photograph. Additionally, **the presence of a glare** is also a detracting factor. This glare creates a visual distraction and diminishes the overall quality of the photograph. The presence of a distracting element in the form of glowing object on the right of the image draws attention away from the main subject, disrupting the composition of the image. In terms of composition, the image **could benefit from a more deliberate framing and a clearer focal point**. The presence of a large number of cubes in the image, while potentially interesting, does not seem to be effectively utilized in this particular photograph. Focusing more on these colored squares results in a more striking composition.* |
| **GPT-4v** | *Aesthetically, the image is quite **appealing**. The colorful geometric shapes hanging from the ceiling create a playful and lively atmosphere. The different sizes and colors of the shapes add depth and visual interest to the space. The reflections on the floor and the soft lighting also contribute to the overall aesthetic of the image. The combination of colors and shapes creates a sense of movement and dynamism, making the space feel energetic and vibrant.* |
| **qwen-vl** | *The image captures a **colorful and artistic** installation in a building's atrium. There are numerous square pieces of paper in different colors hanging from the ceiling, creating a **vibrant and playful atmosphere**. Some of the paper squares are also painted with various colors, adding more diversity to the overall scene. This installation appears to be a creative and engaging artwork, potentially stimulating the viewers' imagination and bringing joy to those who view it.* |
| **spark-multi-3** | *This image shows an interior space with many colorful paper chains hanging from the ceiling. These paper chains present different colors and shapes, creating a **lively, fun atmosphere**. The whole scene gives a relaxed and cheerful feeling that makes one want to explore the space and admire these beautiful decorations.* |
| **cogvlm** | *The picture shows a colorful art installation in the atrium of a shopping mall. There are many multi-colored cubes hanging in the air, and the colors are **bright and vibrant**. These cubes are not the same size, and they are arranged randomly, which makes the whole space look very **dynamic and vibrant**. In addition to the cubes, there is also a small house model placed in the lower right corner of the picture.* |
| **glm-4v** | *This picture has a **good aesthetic** because it uses colorful cubes as decor. These cubes have different colors and sizes, which makes the picture more attractive and interesting. In addition, there are many lights in the background of this photo, which make the colors of the cubes more vibrant and eye-catching. Overall, this image is very **beautiful and creative**.* |

Figure 5: The fifth comparison of aesthetic comments.

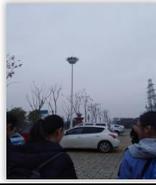

| | |
|---|---|
| **CALM** | *I think this picture could have been **very good if you had gotten closer to your subjects and from a different angle**. Nobody in the picture is really engaging the viewer and there is nothing to hold the viewer's interest.* |
| **CALM-E** | *The image **lacks visual interest** and appears to be a **simple** snapshot of a parking lot. The **composition is unremarkable** and does not draw the viewer in. The **lack of dynamic elements or compelling visual elements** makes the image less appealing. Additionally, the overall light in the image is a bit **dim**, as if it was taken on a cloudy day or in the evening. **Adding light or adjusting the brightness** may be more beneficial to the overall appeal of the picture. Furthermore, the overall aesthetic of the image is quite **dull and uninspiring**. The lack of vibrant colors or interesting textures further contributes to the image's lack of appeal. Overall, the image fails to captivate the viewer and lacks the visual appeal necessary to make it aesthetically pleasing.* |
| **GPT-4v** | *This image presents a glimpse of an overcast day in a busy parking area. The towering street lamp stands prominently against the grey sky, its unique, circular design adding a modern touch to the scene. The leafless trees suggest it could be either autumn or winter. The array of cars parked in the area points towards a place of interest nearby, potentially a park or a public space given the natural surroundings. There is also a noticeable electrical pylon in the background, indicating infrastructure and the presence of power lines, perhaps a reminder of the blend of nature and technology in this environment.* |
| **qwen-vl** | *The image shows a group of people gathered outside on a stone patio near some parked cars. There are a few small trees without leaves and a white lamp post in the background. The sky is gray and overcast, and there's a sense of quiet and calm in the scene.* |
| **spark-multi-3** | *This image shows a city street on a rainy day with a row of buildings in the background. In the picture, we can see two people standing under an umbrella, the color of the umbrella is white and blue. In addition, there is a car parked on the street, surrounded by some trees and telephone poles. The whole scene takes on a **sombre** atmosphere, but also reveals a **sense of serenity and calm**.* |
| **cogvlm** | *This picture shows some people and cars on the road. Specifically, in the top left corner of the picture there is a white street lamp with many eyes on it, and behind the street lamp are several leafless trees. In the middle of the picture are three pedestrians wearing black clothes, two of whom are facing away from the camera, and one is facing towards the camera. There is also a white car parked on the side of the road, and behind the car are two high-voltage towers. On the right side of the picture are two pedestrians wearing blue jackets, but only their backs are visible.* |
| **glm-4v** | *This photo has a **certain aesthetic quality** to it. The bare trees and gray sky create a sense of **melancholy and introspection**, while the girls add **vitality and youthfulness** to the scene. The **contrast** between the cars and the street lamp further enhances the visual interest. Overall, I think this photo is **beautiful and evocative**.* |

Figure 6: The sixth comparison of aesthetic comments.

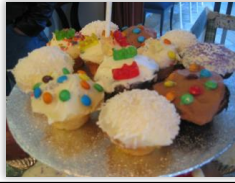

| | |
|---|---|
| **CALM** | *I think the focus is a little off and the background is a little **distracting**. Nobody really seems to be in **focus**.* |
| **CALM-E** | *The cupcakes in this image look **colorful and delicious**, but there are some aesthetic issues about this image that could be improved. The focus of the image is **not sharp**, which detracts from the overall quality. Additionally, the presence of the person in the background is **distracting** and **takes away from the main subject of the cupcakes**. The **composition** of the image could be improved by adjusting the focus and possibly repositioning the cupcakes to create a more visually appealing arrangement. The current layout feels a bit static and could be enhanced by incorporating different angles, perspectives, or props to add visual interest. Additionally, the background can be **blurred** or try a more compact crop may make the cupcakes stand out. Furthermore, the **color balance and saturation** in the image could be adjusted to make the cupcakes appear more vibrant and appetizing. This could involve enhancing the natural colors of the cupcakes and adjusting the overall color balance to create a more visually appealing image. Overall, while the cupcakes themselves are delicious, the technical aspects of the image could be enhanced to make it more visually appealing.* |
| **GPT-4v** | *The cupcakes in this image look **colorful** and inviting with their various toppings such as sprinkles, candies, and coconut flakes. The arrangement on the serving platter is visually **appealing**, with a mix of chocolate and vanilla cupcakes creating a nice contrast. Overall, the cupcakes appear to be a **fun and delightful** treat for any occasion.* |
| **qwen-vl** | *The image showcases an array of **beautifully decorated** cupcakes on a glass plate. The cupcakes are **artistically arranged** in different sizes and positions, each with its unique frosting and topping combinations. The diverse toppings include chocolate, candy, and sprinkles, adding to the visually **appealing** presentation. Overall, the cupcakes' **vibrant colors and attractive arrangement** make for a tempting and eye-catching display.* |
| **spark-multi-3** | *This image shows a silver tray filled with various cupcakes arranged on a white background. Cupcakes come in a variety of flavors and colors, including chocolate, vanilla, and green frosting. The tray sits on a table and looks **delicious and inviting**.* |
| **cogvlm** | *The picture shows a plate of various **delicious** cupcakes. Specifically, there are 10 cupcakes on the plate, and each cupcake is different from the others in terms of color and decoration. There are coconut shavings on some cupcakes, and there are also colorful candies on top of them. In the background is an unidentified person and a chair.* |
| **glm-4v** | *In terms of aesthetics, the image is quite **appealing** due to the **vibrant colors and diverse decorations** on the cupcakes. The variety in toppings, such as gummy bears, candy-coated chocolates, and coconut shavings, adds visual interest and makes the display more tempting. Additionally, the arrangement of the cupcakes on a circular plate enhances their appeal by creating a harmonious balance and showcasing each one individually. Overall, the presentation is both visually **pleasing** and suggests a **festive or celebratory occasion**.* |

Figure 7: The seventh comparison of aesthetic comments.

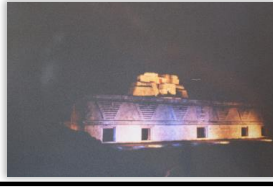

| | |
|---|---|
| **CALM** | *I'm not sure what the big grey smudge is on the left, but it is very **distracting**. Also this is just too **grainy** for me.* |
| **CALM-E** | *The image of the night view of an ancient architecture is quite **captivating**, with the building illuminated against the dark sky. The **contrast** between the bright lights and the darkness of the night creates a visually striking effect. The image of the night shot has a lot of **potential**, but there are a few aesthetic issues that detract from its overall impact. The presence of a large dust spot in the sky is quite distracting and takes away from the overall **composition** of the image. Additionally, **the lack of sharpness** in the image is noticeable and detracts from the overall quality of the shot. In terms of composition, the image **could benefit from a more balanced and visually appealing arrangement of elements**. The placement of the subject and the surrounding elements could be adjusted to create a more dynamic and engaging composition. Additionally, the image is a bit **dimmed** and **details are lost**. **Use of lighting and color could be enhanced** to create a more visually striking and atmospheric image. Overall, while the image has some strong points, such as the interesting subject and strong historical atmosphere, there are several aesthetic issues that detract from its overall impact. With some adjustments to the composition, sharpness, and overall visual appeal, the image has the potential to be a truly stunning night shot.* |
| **GPT-4v** | *This image captures the striking architectural beauty of an ancient structure illuminated at night. The **warm lighting** casts a dramatic glow on the stone façade, highlighting the intricate carvings and geometric patterns that adorn the walls. The overall effect is one of awe and reverence for the skill and craftsmanship of the civilization that constructed this magnificent building.* |
| **qwen-vl** | *The image features an ancient Maya building, possibly a temple or palace, lit up at night with colorful lighting. The building appears to be made of stone and has a rectangular structure with decorative elements on the roof. The lighting creates a **visually striking scene**, emphasizing the architectural details of the building and highlighting its historical significance.* |
| **spark-multi-3** | *This image shows a scene at night where a building is illuminated with blue lights. From an aesthetic point of view, the design of the building is very unique and attractive. Its appearance presents a **futuristic feel**, possibly made of modern materials such as glass or metal. The blue lighting adds a mysterious and dreamy atmosphere to the whole scene, creating a mesmerizing visual effect.* |
| **cogvlm** | *This is a picture of a temple. The temple is made up of yellow and gray bricks, with a dark roof. There are several Windows in the building. The light inside the temple shines out through the Windows, making the temple look **bright**.* |
| **glm-4v** | *This photo is **not very aesthetic** because it is very **dark and lacks contrast**. **The details of the building are vague and hard to see clearly**.* |

Figure 8: The eighth comparison of aesthetic comments.

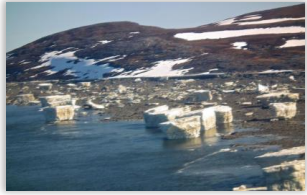

**Image Degradation: Defocus Blur**

| | |
|---|---|
| **CALM** | *A bit **blurry** and the horizon is **tilted**. Otherwise, not bad.* |
| **CALM-E** | *The image captures a beautiful scene of ice and snow, but there are some aesthetic improvements that could be made to enhance the overall impact of the photo. Firstly, the image appears to be slightly **out of focus**, which detracts from the overall sharpness and clarity of the scene. To improve this, **adjusting the focus and ensuring that the subject is clear** and well-defined would greatly enhance the visual appeal of the image. Additionally, the **composition** of the photo could be improved by **straightening the horizon** and ensuring that the main subject is positioned in a way that creates a more visually appealing and balanced composition. This could involve repositioning the camera to capture a more dynamic and engaging perspective of the ice and snow. Furthermore, the **color and contrast** of the image could be enhanced to bring out the natural beauty of the scene. This could involve **adjusting the levels and curves in post-processing** to bring out the richness and depth of the colors, as well as to create a more visually striking and impactful image. Overall, while the subject of the image is interesting and visually appealing, there are several aesthetic improvements that could be made to elevate the overall quality and impact of the photo. By addressing the focus, composition, and color and contrast, the image could be transformed into a more visually stunning and captivating piece of art.* |
| **GPT-4v** | *The image appears to be slightly **blurry** and **out of focus**, which detracts from the crispness and clarity that would enhance the overall aesthetic. Additionally, the **composition** could be improved by adjusting the angle or framing to create a more balanced and harmonious scene. A lower ISO setting and a steadier hand or use of a tripod could help to achieve a sharper and more visually appealing image.* |
| **qwen-vl** | *There is no existence of an aesthetic improvement, but I can suggest a suggestion related to the **content** of the image. If there are indeed snow-covered mountains in the background, it would be interesting to see more details about them, such as the mountain peaks, slopes, or snow patterns. By emphasizing the mountain range, the overall composition of the image would be more captivating and visually appealing, capturing the beauty of the landscape.* |
| **spark-multi-3** | *While the image is already visually appealing, a slight some **color contrast or saturation** to a slight improvement could be to add some color contrast or saturation to certain areas of the image, such as the ice blocks or the sky. This would help draw attention to specific elements and create a more dynamic visual experience for the viewer.* |
| **cogvlm** | ***No**, the image is already very beautiful.* |
| **glm-4v** | *Considering the image's **composition**, one could consider adding a more balanced distribution of icebergs across the foreground and background. This would provide a more harmonious balance in terms of visual weight and depth. Additionally, capturing a time when there was more snow on the hills might enhance the overall mood and beauty of the scene.* |

Figure 9: The first comparison of aesthetic suggestions.

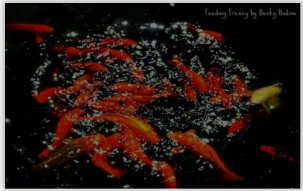

**Image Degradation: Brightness Reduction**

| | |
|---|---|
| **CALM** | *I like the idea, but the image is a little **dark**. I think if it were **brighter**, the colors would stand out more.* |
| **CALM-E** | *The image of the postcard is visually appealing, with vibrant colors and a clear focus on the fish. The composition is well-balanced, and the overall aesthetic is pleasing. However, there are a few aesthetic improvements that could be suggested for this image. Firstly, the overall light in this image is a little too **dim** to effectively highlight the vibrant colors of the fish. The lighting in the image could be adjusted to enhance the colors and details of the fish. This could be achieved through **natural lighting or by using artificial lighting** to create a more dynamic and visually striking effect. Furthermore, the use of a **polarizing filter** could help to reduce glare and enhance the clarity of the image, particularly in the water and the fish. This would result in a more polished and professional-looking photograph. Additionally, considering the use of a **different background** or setting for the fish could add an extra layer of visual interest to the image. Overall, while the image of the postcard is already visually appealing, these suggested aesthetic improvements could further enhance the overall impact and appeal of the photograph.* |
| **GPT-4v** | *The image could use **more contrast** to make the colors of the fish stand out more against background. Additionally, **adding a vignette to the edges** could help draw the viewer's focus towards the center of the image where the fish are located.* |
| **qwen-vl** | *One possible aesthetic improvement for this image could be to **enhance the color balance and contrast** to bring out the vivid colors of the goldfish more effectively. Additionally, **adding a gentle blur or filter** effect to the water around the goldfish could create a soft and soothing look, making the scene more visually appealing.* |
| **spark-multi-3** | *Yes, I would suggest slightly **adjusting the contrast of the image** to make it more visually appealing. For example, you could increase the saturation levels of the colors to bring out their vibrancy. Additionally, using a warmer or cooler tone on some areas could help create a more harmonious overall look.* |
| **cogvlm** | *The image has a few aesthetic flaws that could be improved. Firstly, the fish are swimming in a **murky water** which makes it hard to distinguish between them and their surroundings. Secondly, there is **too much text overlaid on the image** which can distract from the actual content of the photo. Lastly, the **lighting** seems uneven with some parts of the image being brighter than others, causing an imbalance in the overall composition.* |
| **glm-4v** | *Considering the aesthetics of the image, it might be beneficial to include a **more balanced composition** that includes both fish and some surrounding environment. This would give viewers a broader perspective of the setting where these fish live. Additionally, **enhancing the clarity of the water** could make the fish stand out more prominently against their surroundings.* |

Figure 10: The second comparison of aesthetic suggestions.

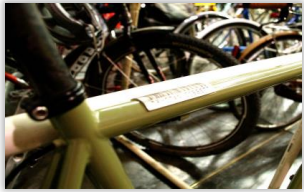

**Image Degradation: Brightness Increase**

| | |
|---|---|
| **CALM** | *Composition here is a mess, nobody will know what the heck it is unless they are bike enthusiasts, and some places are obviously overexposed.* |
| **CALM-E** | *There are a few areas that could be improved to enhance the overall quality of the image. Firstly, the image is a little overexposed in some places, which can detract from the overall visual appeal. Softening the lighting or adjusting the exposure could help create a more balanced and pleasing aesthetic. Additionally, the background of the image is a bit cluttered and distracting. Simplifying the background or using a shallower depth of field to blur the background could help to direct the viewer's focus more effectively. Furthermore, there is a lack of a clear focal point in the image. Adding a focal point, such as a prominent feature or element within the frame, could help to create a more engaging and dynamic composition. Additionally, the overall sharpness and clarity of the image could be improved to enhance the visual impact. Overall, by addressing the lighting, background, focal point, and sharpness, the image could be elevated to a higher level of visual appeal.* |
| **GPT-4v** | *There is a bubble in the protective covering near the Wabi Cycles logo.* |
| **qwen-vl** | *In the image, there are multiple bicycles, and one of them has a yellow and black frame. However, the bicycle has a large scratch on its side, which detracts from its aesthetic appeal. The scratch makes the bicycle appear less appealing and could be a potential drawback when it comes to selling or using the bike as a means of transportation. Addressing this flaw by either repairing or replacing the damaged section or considering a new bicycle with a scratch-free frame would improve the overall appearance and functionality of the bike.* |
| **spark-multi-3** | *The aesthetic flaw I think needs to be improved is the lack of focus on the details of the bicycle. The image is mostly focused on the overall structure of the bike, with little attention paid to the specific parts and components. This could be improved by focusing more on the details of the frame, wheels, brakes, and other important features. Additionally, it would be helpful to have a better understanding of the materials used in the construction of the bike, as well as any unique design elements or customizations that may have been made.* |
| **cogvlm** | *The image shows a green bicycle and a white bar code label on the bicycle, but there is no obvious aesthetic flaw to be improved.* |
| **glm-4v** | *While the image focuses on the barcode and frame of the bike, aesthetically speaking, one might consider adding some protective tape or stickers around the barcode to prevent scratches. Additionally, considering a more visible color for the barcode could enhance its visibility in various lighting conditions.* |

Figure 11: The third comparison of aesthetic suggestions.

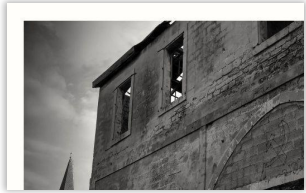

**Image Degradation: Cropping**

| | |
|---|---|
| **CALM** | *I like the B&W color scheme. However, I think the **border is a bit too big**, and I'm not sure how to fix it.* |
| **CALM-E** | *The image of the abandoned building has a haunting and eerie quality to it, which is enhanced by the use of black and white. The contrast between the dark shadows and the lighter areas of the building creates a sense of depth and mystery. However, there are some aesthetic problems that could be addressed to further improve the image. Firstly, the **composition** of the image could be improved. The building is positioned slightly off-center, and there is a meaningless border at the top, which creates a sense of imbalance. **Adjusting the framing to center the building or using the rule of thirds** could create a more visually appealing composition. Additionally, the **lighting** in the image could be enhanced to bring out more details in the building. The shadows are quite deep, and while this adds to the overall mood of the image, it also obscures some of the architectural details. **Adjusting the lighting or using a different time of day** to capture the image could help to bring out more of the building's features. Furthermore, the image could benefit from some **post-processing to enhance the contrast and bring out more details** in the shadows. This could help to create a more dynamic and visually striking image. Overall, while the image effectively captures the haunting and abandoned nature of the building, there are some aesthetic improvements that could be made to enhance the overall impact of the image.* |
| **GPT-4v** | *The building needs to be improved, it needs to be **painted** and has some **broken windows** that needs to be repaired.* |
| **qwen-vl** | *The aesthetic flaw in the image that needs improvement is the **damaged and broken window**. The broken glass and ruins of the window frame give off a negative and aged vibe to the old building, which might be visually unappealing to some viewers. The building, although old, still has architectural beauty that can be appreciated. A restored or repaired window can enhance the overall appearance of the building, giving it a cleaner, more modern look while preserving its historical value.* |
| **spark-multi-3** | *The aesthetic flaw in this image is the **lack of balance between the old and new elements**. The old, dilapidated building contrasts with the modern structure in the background, creating a jarring effect. Additionally, the **cracked and damaged exterior of the building** detracts from its overall appearance. To improve the aesthetic, it may be helpful to add some vegetation or other natural elements to soften the contrast between the old and new.* |
| **cogvlm** | *The wall is **dirty** and needs cleaning.* |
| **glm-4v** | *The image has a monochromatic **color scheme** which can limit the depth and mood of the photograph. Including some color elements could enhance the overall aesthetics and make it more vibrant.* |

Figure 12: The fourth comparison of aesthetic suggestions.

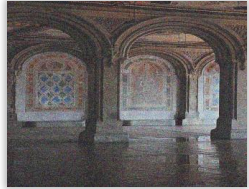

**Image Degradation: Salt-and-pepper Noise**

| | |
|---|---|
| **CALM** | *I'm guessing the **noise** is intentional, but I'm not sure it works here. Good composition though.* |
| **CALM-E** | *The image has a strong composition and the use of light and shadow creates a dramatic effect. The colors are vibrant and the overall aesthetic is visually appealing. However, there is a **noticeable amount of noise** in the image, which detracts from its overall quality. The noise is particularly noticeable in the darker areas of the image and it creates a distraction from the otherwise captivating scene. To improve the aesthetic quality of the image, it would be beneficial to **reduce the noise through post-processing** techniques or by using a **lower ISO** setting when capturing the image. This would help to enhance the overall visual appeal and allow the viewer to fully appreciate the beauty of the scene without being distracted by the noise.* |
| **GPT-4v** | *Image is pixelated and doesn't have very good **resolution**, this can be improved by taking the photo with **more clarity** and **better lighting**.* |
| **qwen-vl** | *In the image, there are three arched doorways in a row with different designs on them. One of the doorways appears to be **older and less well-maintained than the others**, giving an impression of aesthetic flaw. To improve this aspect, it would be helpful to restore the doorway to its original condition, or at least provide a more uniform appearance by cleaning and repainting or retiling the surface. This would enhance the overall visual appeal of the space and create a more harmonious appearance.* |
| **spark-multi-3** | *The aesthetic flaw I think needs to be improved is the **lack of lighting** in the image. The dark shadows and unclear details make it difficult to see the details of the architecture and decorations. Additionally, the overall feeling of the image could be enhanced by adding some **brighter colors** or highlights to highlight certain features.* |
| **cogvlm** | *The image shows a large and spacious hall with high ceilings, but the **light is dim**. It seems that there are some architectural decorations on the walls and ceiling, but it is difficult to see clearly because of the poor lighting. In my opinion, improving the lighting would make the space look more beautiful and vibrant.* |
| **glm-4v** | *The image seems to have a slightly **grainy quality**, which might be due to the camera settings or the resolution of the image. This could be improved by using a **higher-resolution** camera or post-processing techniques to enhance clarity.* |

Figure 13: The fifth comparison of aesthetic suggestions.

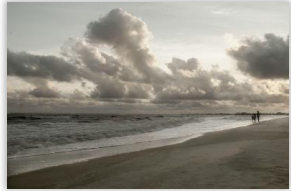

**Image Degradation: Color Saturation Reduction**

| | |
|---|---|
| **CALM** | *Nice. The clouds and the water are **so close in color**. I might have done this in B&W to give it a little more punch. Still a nice photo.* |
| **CALM-E** | *The composition of the image is quite striking, with the dramatic clouds and the silhouettes of the people on the beach creating a sense of depth and scale. However, the overall aesthetic of the image could be improved in a few ways. Firstly, **the color balance and saturation could be adjusted** to make the image more vibrant and dynamic. This could help to bring out the natural beauty of the scene and make it more visually appealing. The contrast between the dark silhouettes and the bright sky could also be enhanced to make the image more visually impactful. This could be achieved through **adjustments in post-processing to bring out more detail** in the silhouettes and make the sky appear even more dramatic. Additionally, the overall **sharpness and clarity of the image** could be improved. The details in the clouds and the texture of the beach could be brought out more to create a more immersive and engaging visual experience for the viewer. Overall, while the composition and subject matter of the image are strong, there are several aesthetic improvements that could be made to enhance the overall impact and visual appeal of the photograph.* |
| **GPT-4v** | *The portrayal could be a bit **dark**, just a little light need to be added.* |
| **qwen-vl** | *There is **no** existence of an aesthetic flaw in the image.* |
| **spark-multi-3** | *The aesthetic flaw I think needs to be improved is the **lack of contrast** in the image. The sky and water are both very light, making it difficult to see details in the image. Additionally, the buildings and small islands in the distance are not clearly visible due to their low contrast with the background.* |
| **cogvlm** | *The image **quality** is a bit low, and some details are not very clear. It would be better if the picture was clearer with more distinct features.* |
| **glm-4v** | *The main aesthetic flaw in this image is the presence of a **large amount of clouds**, which obscure the view and make it difficult to see clearly.* |

Figure 14: The sixth comparison of aesthetic suggestions.

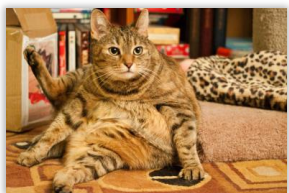

**No Image Degradation**

| | |
|---|---|
| **CALM** | *I like the expression on the cat's face, but the **background is a bit distracting**.* |
| **CALM-E** | *The image of the cat is indeed very cute and the pose is well-captured. The cat's expression is endearing and the overall composition of the image is visually appealing. However, there are some aesthetic problems that could be addressed to enhance the overall quality of the image. Firstly, the **background of the image** is quite busy and distracting. The cluttered background takes away from the focus on the cat and makes the image feel a bit chaotic. It would be beneficial to either **simplify the background or use a shallower depth of field** to blur out the distracting elements. Additionally, the **color scheme** of the picture can be further improved Both the background and the cat are warm yellow, making the main body of the image less prominent. Consider using some **post-processing techniques can make the color contrast** more prominent. Furthermore, while the cat's pose is indeed cute, the **composition** of the image could be improved. The framing of the cat within the image feels a bit off-center, and there is a lack of balance in the composition. **Adjusting the framing and composition** to create a more visually pleasing arrangement would greatly enhance the overall aesthetic of the image. Overall, while the image of the cat is undeniably cute, there are some aesthetic problems that could be addressed to improve the overall quality of the image. By addressing the background, lighting, and composition, the image could be elevated to a higher level of visual appeal.* |
| **GPT-4v** | *The arm of the cat is the most distracting feature in this photo. It appears that the cat is **raising its arm** in an awkward manner, which detracts from the overall aesthetic of the image. Additionally, the **background is cluttered** with a variety of items, such as books and a cardboard box, which adds to the visual distraction. To improve the aesthetic, the arm position could be adjusted, and the background could be cleaned up or blurred to create a more cohesive and visually appealing image.* |
| **qwen-vl** | *The image features an aesthetic flaw of **a fat cat**, which is a concern as it may indicate health problems. A proper diet and regular exercise are essential for the health and well-being of cats. This is particularly important as obesity in cats is associated with a range of health issues, such as diabetes, joint problems, heart disease, and shortened life expectancy. Therefore, improving the image by highlighting the issue of obesity in cats could be a step towards promoting responsible pet ownership and encouraging cat owners to make positive changes in their pets' lifestyles.* |
| **spark-multi-3** | *The aesthetic flaw I think needs to be improved is the **lack of depth** in the image. The background is flat and lacks texture, which makes it difficult to focus on the details of the scene. Additionally, there are several elements that could be moved or repositioned to create a more balanced composition.* |
| **cogvlm** | *The cat's body is a little **chubby**, so it needs some exercise.* |
| **glm-4v** | *The image does **not** have any discernible flaws or areas that require improvement. It is a clear and well-composed photograph of a cat in an indoor setting.* |

Figure 15: The seventh comparison of aesthetic suggestions.

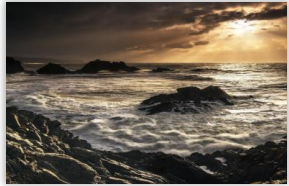

**No Image Degradation**

| | |
|---|---|
| **CALM** | *I love the lighting in this photo. The light beams really draw the eye to the rocks.* |
| **CALM-E** | *The image is truly captivating. The contrast between the dark, moody sky and the bright, sunlit water creates a stunning visual impact. The composition is well-balanced, with the rocks in the foreground adding depth and texture to the scene. The play of light and shadow is masterfully captured, adding to the overall dramatic effect of the image. To make this picture even more attractive, consider enhancing the colors to make them more vibrant and eye-catching. Additionally, experimenting with different angles and perspectives could add an extra layer of visual interest to the composition. Perhaps capturing the scene from a lower angle or incorporating some elements of the surrounding environment could provide a fresh and unique take on the subject matter. Overall, this image is already quite impressive, but with some minor adjustments and creative experimentation, it has the potential to become even more visually striking and engaging.* |
| **GPT-4v** | *The picture seems overly bright and could be toned down a bit with the use of a polarizing filter or by adjusting the exposure settings.* |
| **qwen-vl** | *The image appears to be overly dark, with little contrast between the sky, sea, and land. The clouds, water, and rocks all appear to be in shades of gray and brown, giving the impression of a somewhat dreary and monotonous landscape. To improve the image, additional color and contrast could be introduced, such as lighting up the sky with different colors or enhancing the dark areas to reveal more details in the water and rocks. This would create a more visually appealing and dynamic scene.* |
| **spark-multi-3** | *The colors in the image are a bit off. The sky is too orange and the sun is too bright.* |
| **cogvlm** | *The waves are too strong and rough, which may cause people to feel uncomfortable or fearful.* |
| **glm-4v** | *The contrast between the dark and light areas could be improved. This can be achieved by adjusting the exposure settings to balance out the brightness and shadows more effectively.* |

Figure 16: The eighth comparison of aesthetic suggestions.



\end{document}